%% file: main.tex
\documentclass[10pt]{article} 
\usepackage[accepted]{tmlr}

\usepackage{microtype}
\usepackage{graphicx}
\usepackage{subcaption}
\usepackage{booktabs} 

\usepackage{amsmath}
\usepackage{amssymb}
\usepackage{mathtools}
\usepackage{amsthm}
\usepackage{multirow}
\usepackage{placeins}
\usepackage{wrapfig}

\input{math_commands.tex}

\usepackage{hyperref}
\usepackage{url}

\usepackage[capitalize,noabbrev]{cleveref}

\theoremstyle{plain}

\theoremstyle{definition}

\theoremstyle{remark}

\usepackage{algorithm}
\usepackage{algpseudocode}

\definecolor{updatecolor}{rgb}{0., 0.08, 0.8}

\title{Test-Time Scaling of Text-to-Image Flow Matching Models \\ with Noise-Rejuvenated Sequential Monte Carlo}

\author{\name Wenyi Lian \email wenyi.lian@it.uu.se \\
      \addr Department of Information Technology\\
      Uppsala University, Sweden
      \AND
      \name Zheng Zhao \email zheng.zhao@liu.se \\
      \addr Department of Computer and Information Science \\
      Link\"{o}ping University, Sweden
      \AND
      \name Thomas B. Sch{\"o}n \email thomas.schon@uu.se\\
      \addr Department of Information Technology\\
      Uppsala University, Sweden}

\def\month{10}  
\def\year{2026} 
\def\openreview{\url{https://openreview.net/forum?id=HoQBaPXhKk}} 

\begin{document}

\maketitle

\input{sec/0-abstract}
\input{sec/1-introduction}
\input{sec/2-background}
\input{sec/3-method}
\input{sec/3.1-related_work}
\input{sec/4-experiment}

\input{sec/5-discussion}

\input{sec/6-conclusion}




\bibliography{main}
\bibliographystyle{tmlr}

\appendix

\input{sec/X-appendix}

\end{document}

%% file: math_commands.tex
\usepackage{amsmath,amsfonts,bm}

\def\eqref#1{equation~\ref{#1}}
\def\Eqref#1{Equation~\ref{#1}}

\def\1{\bm{1}}

\DeclareMathAlphabet{\mathsfit}{\encodingdefault}{\sfdefault}{m}{sl}
\SetMathAlphabet{\mathsfit}{bold}{\encodingdefault}{\sfdefault}{bx}{n}



%% file: sec/0-abstract.tex
\begin{abstract}
Test-time scaling of diffusion models with sequential Monte Carlo (SMC) has become a promising strategy for sampling under user-specified rewards, such as human preference scores. However, the standard SMC framework typically relies on stochastic proposals and therefore does not directly transfer to flow matching models, in which inference is governed by deterministic ODEs. In this work, we introduce noise-rejuvenated SMC, an inference-time scaling method tailored for text-to-image flow matching models. Our key idea is to perturb each resampled particle by changing its source noise and constructing a new intermediate state using its current endpoint estimate. This moves the particle to another state that remains representable under the flow-matching interpolation form, enabling particle exploration without converting the sampler to an SDE. We further introduce a look-ahead mechanism to refine resampled particles before weighting. Experiments on Stable Diffusion 3, FLUX.1-dev, and Z-Image demonstrate consistent improvements in reward-guided generation, while ablation studies validate the contributions of noise rejuvenation and look-ahead refinement. Code is available at \url{https://github.com/shermanlian/NR-SMC}.
\end{abstract}

%% file: sec/1-introduction.tex
\section{Introduction}

Inference-time scaling of diffusion models has shown promising results by tilting the stochastic generation towards user-specified target distributions~\citep{ben2024d,eyring2024reno,ma2025inference,singhal2025general} using explicitly defined reward signals such as the CLIP-score~\citep{radford2021learning} and human preference scores~\citep[HPS,][]{xu2023imagereward,wu2023human}. Despite their strong performance, many prior methods~\citep{ben2024d,wang2025training,qi2024not} require backpropagating reward gradients during inference and therefore assume differentiable reward functions. This increases inference cost and limits their applicability to large-scale models and complex or black-box reward signals.

Sequential Monte Carlo (SMC) methods offer a natural alternative by using reward evaluations only for reweighting and resampling particles, rather than for gradient-based optimization~\citep{wu2023practical,Zhao2024rsta,Zhao2025B0SMC,singhal2025general}. As a result, SMC can guide diffusion sampling without differentiating through the reward function or performing test-time training. By maintaining multiple trajectories, SMC enables exploration and progressively twists the sampling process toward regions with higher user-specified rewards~\citep{singhal2025general,ma2025inference}. At the same time, standard SMC algorithms rely on stochastic sampling processes, and are therefore not directly applicable to flow matching (FM) models~\citep{lipman2022flow,liu2022flow}, , which have become an important class of generative models.

\begin{figure}[t]
  \begin{center}
    \centerline{\includegraphics[width=0.95\linewidth]{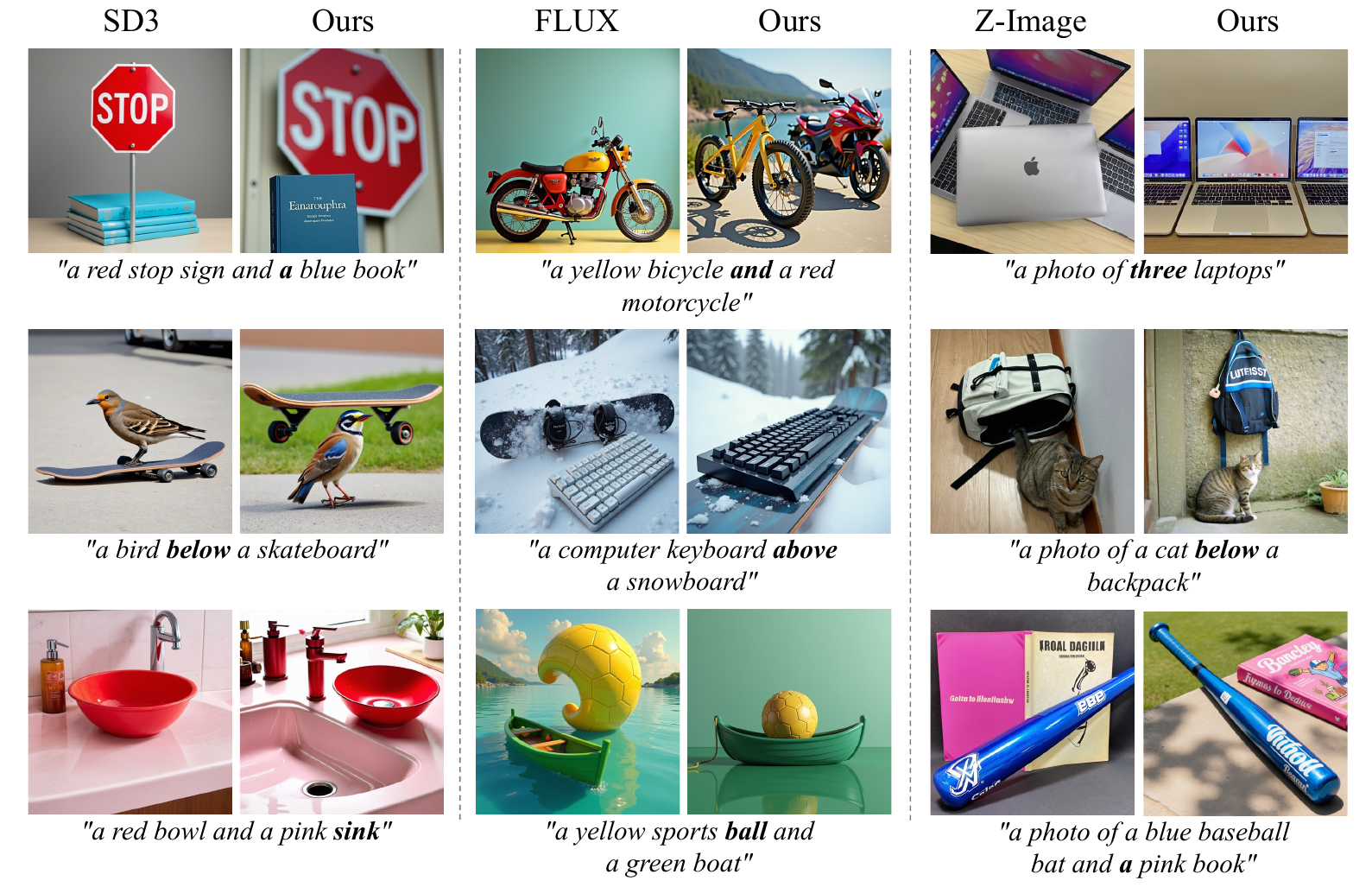}}
    \caption{Qualitative comparison between vanilla sampling and our method on text-to-image generation using Stable Diffusion 3 (SD3)~\citep{esser2024scaling}, FLUX.1~\citep{labs2025flux}, and Z-Image~\citep{team2025zimage}. Our method steers generation toward higher reward objectives, producing outputs that better align with the input prompts without additional training or gradient computation.}
    \label{fig:teaser}
  \end{center}
  \vskip -0.2in
\end{figure}

To obtain the stochastic transitions needed by SMC in deterministic FM samplers, a straightforward baseline is to perturb particle states directly, for example by injecting noise at each solver step. While this does introduce additional stochasticity, it also deviates from the flow matching ODE-based integration path:
\begin{equation}
    z_1 = z_0 + \int_0^1 v_\theta(z_t, t)dt,
    \label{eq:fm-ode}
\end{equation}
where $z_0$ and $z_1$ denote the source and target samples, respectively, and $v_\theta(z_t, t)={dz_t}/{dt}$ is the learned velocity field~\citep{liu2022flow}. In this scenario, adding noise to~\cref{eq:fm-ode} often degrades the generation quality and requires more sampling steps for denoising~\citep{liu2022flow}. As a consequence, most existing SMC approaches fail to improve the results in flow matching models, since they do not generate new paths.

To address this issue, we introduce noise-rejuvenated SMC, a resampling-time perturbation strategy for ODE-based flow matching samplers. Our method keeps the original flow matching solver for normal propagation and injects stochasticity only when particles are resampled. Specifically, after computing the current endpoint estimate from the current particle, we replace its source noise with newly sampled Gaussian noise and construct a rejuvenated intermediate state using the flow-matching interpolation parameterization. The resulting state is subsequently propagated using the original flow-matching solver. This operation constructs an alternative intermediate state compatible with the flow-matching interpolation parameterization, providing controlled exploration while avoiding the quality degradation caused by injecting noise at every solver step.
We further introduce a look-ahead mechanism that refines the rejuvenated particles with an additional flow-matching update, leading to more informative resampling decisions. 

We evaluate our approach on several state-of-the-art flow matching text-to-image models, including Stable Diffusion 3 (SD3)~\citep{esser2024scaling}, Flux.1~\citep{labs2025flux}, and Z-Image~\citep{team2025zimage}. The results show that our method effectively steers generation toward user-specified reward objectives at inference time, without requiring additional training or gradient-based optimization. \autoref{fig:teaser} illustrates representative examples comparing vanilla sampling and our approach.

In summary, our contributions are as follows:
\begin{itemize}
    \item We present a novel inference-time scaling method for flow matching based on sequential Monte Carlo. Our method enables particle exploration for diversity while preserving generation fidelity.
    \item We introduce a resampling-time perturbation strategy, referred to as noise rejuvenation, which introduces stochasticity over the ODE path and thus unleashes the potential of SMC for flow matching applications.
    \item We integrate a look-ahead mechanism allowing for further refinement of the resampled particles.
    \item We empirically show that our proposed method is effective across various state-of-the-art text-to-image flow matching models.
\end{itemize}

%% file: sec/2-background.tex
\section{Background}

\subsection{Diffusion Models}

Diffusion models~\citep{sohl2015deep,ho2020denoising}, in the context of generative modeling, constitute a forward noising process and a reverse process for new data sampling. Specifically, diffusion models can be characterized by the following forward-time stochastic differential equation (SDE):
\begin{equation}
    dz_t = f(z_t, t)dt + g(t)dw_t,
\end{equation}
where $f$ is the drift function, $g$ is the diffusion function, and $w_t$ is a standard Wiener process. Its reverse-time process is given by~\citep{sohl2015deep}
\begin{equation}
    dz_t = [f(z_t, t) - g(t)^2 \nabla \log p_t(z_t)]dt + g(t)d\bar{w}_t,
\end{equation}
where $\nabla \log p_t(z)$ is the score function and $\bar{w}_t$ is a reverse-time Wiener process.

\subsection{Flow Matching}

Flow matching models~\citep{liu2022flow,lipman2022flow} are ODEs defined for time $t \in [0,1]$: $d z_t = v(z_t, t)dt$, where $v(z_t, t)$ is a velocity field that drives~$z_0$ to~$z_1$ following the ODE integration path in~\Cref{eq:fm-ode}. Here, we choose the velocity field $v(z_t, t)$ to be the minimizer of
\begin{equation}
    \min_v \int_0^1 \mathbb{E} [\| (z_1 - z_0) - v(z_t, t) \|_2^2] dt,
\end{equation}
where $z_t$ is a linear interpolation of $z_0$ and $z_1$:
\begin{equation}
    z_t = tz_1 + (1-t) z_0.
    \label{eq:linear_interpolation}
\end{equation}
During inference, we often discretize the solution by
\begin{equation}
    z_{t_{k+1}} = z_{t_k} + \int^{t_{k+1}}_{t_k} v_\theta(z_{s}, {s}) \, ds \approx  z_{t_k} + v_\theta(z_{t_{k}}, t_k) \Delta_k,
\end{equation}
where $\Delta_k = t_{k+1} - t_k$ and $\theta$ denotes the learned parameters of a neural network. For notational simplicity, we identify the discretized time index $t_k$ with $t \in [0,1]$ and use this convention throughout the remainder of the paper.

%% file: sec/3-method.tex
\section{Method}
\label{sec:method}

In this work, we consider flow matching in the latent space for conditional image generation. Specifically, let us denote the image by~$x$ and adopt a decoder $x = D_\phi(z)$ through e.g., a variational auto-encoder~\citep[VAE,][]{kingma2013auto,rombach2022high} with pretrained parameters~$\phi$. The latent space is governed by a flow matching model
\begin{equation}
    z_1 = z_0 + \int_0^1 v_\theta(z_t, t, c)\, dt, \quad z_0 \sim \mathcal{N}(0,I),
    \label{eq:fm-t2i}
\end{equation}
where $c$ denotes the condition prompt, and $v_\theta(z_t, t, c)$ is a prompt-conditional velocity field parameterized by a neural network (often a U-Net~\citep{ronneberger2015u} or a Transformer~\citep{peebles2023scalable}). During inference, sampling starts from an initial Gaussian noise $z_0$ and follows the learned ODE dynamics toward the final latent state $z_1$ at $t=1$. The final image is then obtained by decoding the final $z_1$ as $x=D_\phi(z_1)$.

\subsection{Reward-Guided Conditional Distribution}
One way to steer prompt-conditional generation is to reward-tilt the pretrained conditional distribution under a fixed prompt. Throughout this section, $c$ denotes the same text prompt used by both the flow-matching sampler and the reward model: the velocity field $v_\theta(z_t,t,c)$ induces the base prompt-conditional distribution $p_\theta(z_1\mid c)$, while $r(x,c)$ is an external scalar score that evaluates the decoded image $x$ under the same prompt. Formally, we aim to sample from the reward-tilted conditional distribution:
\begin{equation}
    p_r(z_1 \mid c) \propto p_\theta(z_1 \mid c) \exp \left(\lambda \cdot r(D_\phi(z_1), c)\right),
    \label{eq:reward_guided_distribution}
\end{equation}
where $p_\theta$ is the pretrained generative model distribution, and $\lambda$ controls the strength of the reward tilt. When $\lambda=0$, this distribution reduces to the original prompt-conditional model. The reward model can take various forms, such as vision-language alignment models~\citep{radford2021learning,li2022blip} and human preference scores~\citep{xu2023imagereward,wu2023human}. Sampling from \Cref{eq:reward_guided_distribution} can be performed using particle-based methods, such as sequential Monte Carlo, where the distribution $p_r(z_1 \mid c)$ is approximated by a set of weighted particles~\citep{doucet2001sequential}.

Our work focuses on SMC samplers. More specifically, we maintain a set of interacting particles, each of which represents a flow matching trajectory. Analogously to \citet{singhal2025general}, we leverage a Feynman--Kac model and SMC samplers as the base framework and define a trajectory potential function $G_t(z_{0:t}, c)$ for assigning weights to particles, which is defined as
\begin{equation}
    G_t(z_{0:t}, c) = \exp( \lambda \cdot \max \{ \bar{r}(z_s, c) \}_{s=0}^t),
    \label{eq:patial_reward}
\end{equation}
where $\bar{r}(z_t, c) = r(D_{\phi}(\hat{z}_{1,t}), c)$, and $\hat{z}_{1,t}$ is a flow matching endpoint estimate at time $t$, which can be obtained from the learned velocity field:
\begin{equation}
    \hat{z}_{1,t} = z_t + (1 - t) v_\theta(z_t, t, c).
    \label{eq:z1_estimation}
\end{equation}
See Appendix~A for more details. Then our particles aim to maximize the reward-guided conditional likelihood in \Cref{eq:reward_guided_distribution} over the whole generation trajectory starting at time~$t$ and ending at time~$1$, according to
\begin{equation}
    p_{r,t}(z_{0:t} \mid c) \,
    \propto \, p_\theta(z_{0:t} \mid c) \prod_{s=0}^t \, G_s(z_{0:s}, c) = \, \underbrace{p_\theta(z_t \mid z_{t-1}, c) \, G_t(z_{0:t}, c)}_{\text{Tilted transition kernel}}  \times \; \underbrace{p_\theta(z_{0:t-1} \mid c) \prod_{s=0}^{t-1} \, G_s(z_{0:s}, c)}_{\propto \, p_{r, t-1}}.
    \label{eq:kf_obj}
\end{equation}
This recursive equation pushes the weighted path $z_{0:t}$ to yield high-reward latent samples $z_1$. In addition, we define $G_1 = \exp(\lambda \cdot r(x(z_1), c) (\prod_{t=0}^{1-\Delta t} G_t)^{-1}$ (where $\Delta t$ is the time interval) and set the terminal accumulated potentials to match the exponential tilt term in \Cref{eq:reward_guided_distribution} i.e., $\prod_{t=0}^1 G_t = \exp(\lambda \cdot r(x(z_1), c))$, such that
\begin{equation}
    p_{r,t}(z_{0:1} \mid c) \propto p_\theta(z_{0:1} \mid c) \exp(\lambda \cdot r(x(z_1), c)).
\end{equation}
By letting the proposal distribution of SMC exactly be the model's transition kernel $p_\theta(z_t \mid z_{t-1}, c)$, the weighting function is given by the potential $G_t(z_{0:t}, c)$. We then resample particles proportionally to their weights, computed from the potential, which gradually tilts the generation toward high-reward distributions.

\subsection{Noise Rejuvenation for Flow Matching}
\begin{figure}[t]
   \begin{center}
     \centerline{\includegraphics[width=0.75\linewidth]{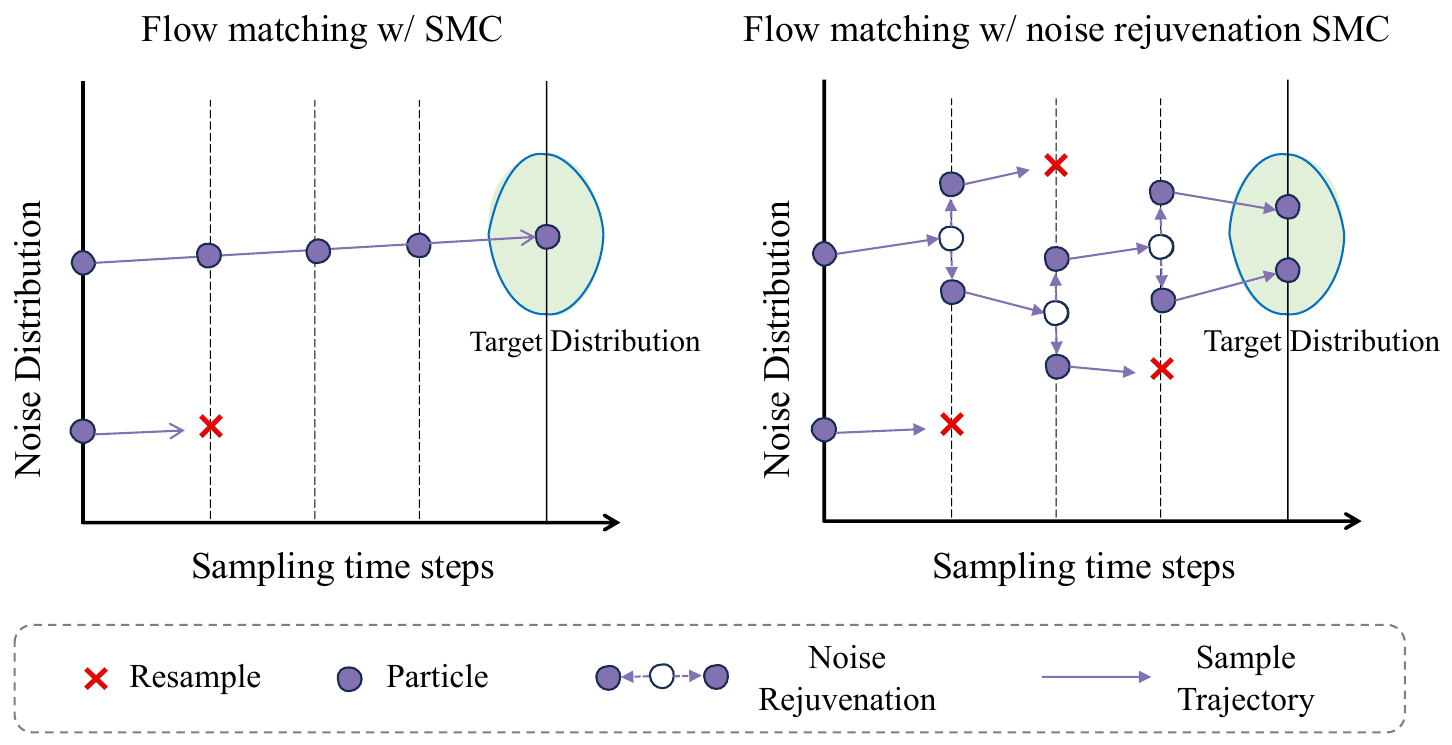}}
     \caption{Illustration of flow matching using standard SMC (left) and our noise-rejuvenated SMC (right) during inference-time sampling. For clarity, we only visualize two particles (purple). When directly applying SMC to flow matching models, resampling leads to particle collapse, producing identical trajectories and final outputs (left). In contrast, noise rejuvenation injects controlled stochasticity during resampling, inducing occasional ``jumps” that allow particles to escape collapsed trajectories and explore alternative paths throughout the sampling process (right).} 
     \label{fig:method}
   \end{center}
   \vskip -0.2in
 \end{figure}

While flow matching employs ODE solvers as the default choice for high-quality data generation, the resulting deterministic dynamics are not directly compatible with traditional SMC methods, which rely on stochastic transitions to maintain particle diversity during sampling. See ~\autoref{fig:method} for an illustration. 

One possible remedy is to convert the FM sampler into an SDE or inject state noise at every solver step. We avoid this approach because it replaces the default ODE inference path with a globally stochastic sampler, introduces an additional noise schedule, and injects randomness even when no resampling-induced collapse occurs. Such global stochasticity may move particles away from the learned FM integration path and degrade generation fidelity. We instead introduce noise rejuvenation, which injects stochasticity only at resampling steps to mitigate particle collapse while largely retaining the behavior of the underlying ODE sampler.

Formally, given state $z_t$, initial noise $z_0$, and target estimate $\hat{z}_{1,t}$ at time $t$, the original interpolation is defined in \Cref{eq:linear_interpolation}. We first compute the pre-rejuvenation endpoint estimate $\hat{z}_{1,t}$ using~\Eqref{eq:z1_estimation}. Then, during the rejuvenation update, this estimate is held fixed while the source noise $z_0$ is replaced with newly sampled Gaussian noise:
\begin{equation}
    \tilde{z}_t = (1-t)\tilde{z}_0 + t \hat{z}_{1,t},
    \label{eq:noise_rejuvenation}
\end{equation}
where $\tilde{z}_0 \sim \mathcal{N}(0, I)$ and $\hat z_{1,t}=z_t+(1-t)v_\theta(z_t,t,c)$. This noising interpolation explicitly constructs a Gaussian corruption kernel for state $z_t$:
\begin{equation}
    \begin{aligned}
\pi_\theta(\tilde z_t|z_t)
=
\mathcal N(\tilde z_t; t\hat z_{1,t}, (1-t)^2I).
    \label{eq:nr_kernel}
    \end{aligned}
\end{equation}
The variance of the rejuvenation kernel decreases as
$t\rightarrow1$, so the magnitude of the perturbation becomes smaller near the end of the sampling process. Then we can rewrite \Cref{eq:kf_obj} as
\begin{equation}
    \tilde{p}_{r,t}(\tilde{z}_{0:t} \mid c) \, \propto \, \tilde{p}_\theta(\tilde{z}_{0:t} \mid c) \prod_{s=0}^t \, G_s(\tilde{z}_{0:s}, c) = \, \tilde{p}_\theta(\tilde{z}_t \mid \tilde{z}_{t-1}, c) \, G_t(\tilde{z}_{0:t}, c) \times \tilde{p}_\theta(\tilde{z}_{0:t-1} \mid c) \prod_{s=0}^{t-1} \, G_s(\tilde{z}_{0:s}, c),
\label{eq:nr_obj}
\end{equation}
where the rejuvenated transition kernel is
\begin{equation}
    \tilde{p}_\theta(\tilde{z}_t \mid \tilde{z}_{t-1}, c) = \int \pi_\theta(\tilde{z}_t\mid z_t) \, p_\theta(z_t \mid \tilde{z}_{t-1}, c) dz_t.
    \label{eq:nr_transition}
\end{equation}
From an intuitive perspective, the propagation can be viewed as a noise rejuvenation step followed by forward flow matching. 
Let $\tilde{p}_{\theta}(\tilde{z}_1 \mid c)$ denote the terminal marginal induced by the rejuvenated transitions. The resulting procedure can then be interpreted as targeting the reward-tilted distribution:
\begin{equation}
    \tilde{p}_r(\tilde{z}_1 \mid c)\propto\tilde{p}_{\theta}(\tilde{z}_1 \mid c)\exp(\lambda \cdot r(x(\tilde{z}_1), c)).
    \label{eq:nr_guided_distribution}
\end{equation}
The above equation characterizes the reward-guided distribution associated with the rejuvenated dynamics. The rejuvenated transition kernel is then incorporated into the standard SMC weighting and resampling procedure, where the potentials $G_t$ are used to assign particle weights. Each rejuvenation step constructs an alternative intermediate state using a resampled source noise and the fixed pre-rejuvenation endpoint estimate, allowing resampled particles to explore different subsequent trajectories. The complete procedure is summarized in Algorithm~\ref{alg:nr_smc}.

\begin{algorithm}[t]
  \caption{Noise Rejuvenation SMC for Flow Matching}
  \label{alg:nr_smc}
  \begin{algorithmic}
    \State{\bfseries Input:} Prompt $c$, flow matching model $v_\theta(z_{t}, t, c)$, VAE decoder $\mathcal{D}_\phi$, potentials $G_t$, number of particles $N$
    \State {\bfseries Initialize:} $\tilde{z}_0^{i} \sim \mathcal{N}(0, I)$, weights $w_0^{i}=1/N$
    \For {$t=0, \dotsc, T-1$}

    \State \textbf{Resample:} Draw indices $a_t^i \sim \mathrm{Multinomial}(w_t)$ and form the resampled particle pairs $\{(\tilde{z}_t^{a_t^i}, w_t^{a_t^i})\}_{i=1}^N$.
    \State {\bfseries Propagate:} Draw $\tilde{z}_{t+1} \sim \tilde{p}_\theta(\tilde{z}_{t+1} | \tilde{z}_{t}, c)$ from \Eqref{eq:nr_transition}.
    \State {\bfseries Re-weight:} Compute the weights $\tilde{w}_{t+1}^i=G_{t+1}(\tilde{z}_{0:t+1}, c)$, and normalize $w_{t+1}^i=\frac{\tilde{w}_{t+1}^i}{\sum_{j=1}^N \tilde{w}_{t+1}^j}$.
    \EndFor
    \State {\bfseries Output:} return samples and weights $\{ \mathcal{D}_\phi(\tilde{z}_1^i), \, w_T^i\}_{i=1}^N$
    
  \end{algorithmic}
\end{algorithm}

\subsection{Look-ahead Refinement}
\label{subsec:LA}

Noise rejuvenation changes the particle state immediately after resampling, as illustrated in \autoref{fig:method} (right). Although this operation preserves particle diversity, the reward evaluated immediately after rejuvenation may be less reliable because the perturbed particle has not yet been propagated by the flow matching dynamics. To obtain a more informative and stable estimate for particle selection, we introduce a one-step look-ahead refinement.
Concretely, after the noise rejuvenation produces a perturbed state $\tilde z_t$, we apply one additional deterministic flow matching update
\begin{equation}
    \tilde z_{t+\Delta t}^{\mathrm{LA}} = \tilde z_t + \Delta t \cdot v_\theta(\tilde z_t,t,c),
\end{equation}
where $\Delta t$ is the solver step size. We utilize \Eqref{eq:z1_estimation} to estimate the corresponding terminal latent from this look-ahead state as
\begin{equation}
    \hat z_{1,t}^{\mathrm{LA}} = \tilde z_{t+\Delta t}^{\mathrm{LA}} + (1-t-\Delta t) v_\theta(\tilde z_{t+\Delta t}^{\mathrm{LA}}, t+\Delta t, c).
\end{equation}
We then compute its reward $r_t^{\mathrm{LA}}$ and define a look-ahead potential according to
\begin{equation}
    G_t^{\mathrm{LA}}(z_{0:t},c) = \exp\left(\lambda \max \left[\{ \bar r(z_s,c) \}_{s=0}^{t}, r_t^{\mathrm{LA}} \right] \right).
\end{equation}
This potential evaluates each rejuvenated particle after a short roll-out under the same flow matching dynamics. Therefore, resampling is guided not only by the current perturbed state, but also by an estimate of how the particle will evolve in the next step. From the perspective of SMC, this can be viewed as an auxiliary potential that approximates the future contribution of each particle to the terminal reward, leading to more stable and informative particle selection.

%% file: sec/3.1-related_work.tex
\section{Related Work}

Recent studies in text-to-image generation have shown that prompt adherence is a key challenge for diffusion-based models~\citep{xu2023imagereward,chefer2023attend,chen2024training,feng2022training,feng2023layoutgpt}, for example, missing objects, incorrectly bound attributes, and wrong locations~\citep{ghosh2023geneval}. Here, we focus on addressing these problems using inference-time scaling techniques, such as gradient-based optimization~\citep{clark2023directly,prabhudesai2023aligning} and gradient-free guidance~\citep{kim2025test,jain2026diffusion}. Particularly, gradient-free methods based on SMC have drawn increasing attention since they require less computation and alleviate over-optimization~\citep{liu2023correcting,singhal2025general}. 

SMC~\citep{doucet2001sequential,naesseth2019elements} has emerged as a versatile framework for probabilistic modelling, providing effective tools for sampling and inference~\citep{johansen2009tutorial,ryan2016review,hol2006resampling,lindsten2014particle}. More recently, SMC has been applied to autoregressive~\citep{zhao2024probabilistic,loula2025syntactic,dang2025inference} and diffusion models~\citep{jasra2009sequential,kim2025alignment,wu2025reverse,chen2024diffusion,skreta2025feynman} as a powerful sampler capable of drawing data from user-preferred distributions. In particular, SMC methods can be perfectly integrated into the inference of pretrained diffusion models, in which external reward signals, such as CLIP scores~\citep{radford2021learning} and Human preference scores~\citep{wu2023human,xu2023imagereward}, are used to guide the diffusion process towards high-reward target areas. \citet{singhal2025general} formulate this as a Feynman-Kac interacting particle system. Subsequent methods further introduce alternative sampling strategies, such as Gibbs sampling or importance sampling, to improve particle diversity and efficiency~\citep{ou2025inference,dang2025inference,liu2023correcting}. However, these methods are primarily designed for stochastic diffusion models and rely on stochastic proposals or SDE-based sampling dynamics.

%% file: sec/4-experiment.tex
\section{Experiments}

\begin{table}[t]
\caption{Text-to-image comparison on the GenEval~\citep{ghosh2023geneval} benchmark across different pretrained backbones. We report GenEval, ImageReward~\citep{xu2023imagereward}, HPSv2~\citep{wu2023human}, and Correct Image~\citep{ghosh2023geneval}. Bold numbers indicate the best result under each backbone.}
\label{tab:overall}
\centering
\begin{small}
\resizebox{0.78\linewidth}{!}{
\begin{tabular}{llcccc}
\toprule
Backbone & Method & GenEval & ImageReward & HPSv2 & Correct Image \\
\midrule
\multirow{3}{*}{SD3}
& Vanilla      & 0.718 & 1.081 & 0.302 & 0.707 \\
& FK-Diffusion & 0.636 & 1.080 & 0.293 & 0.624 \\
& Ours         & \textbf{0.748} & \textbf{1.268} & \textbf{0.309} & \textbf{0.738} \\
\midrule
\multirow{3}{*}{FLUX.1-dev}
& Vanilla      & 0.663 & 0.986 & 0.306 & 0.646 \\
& FK-Diffusion & 0.467 & 0.953 & 0.288 & 0.450 \\
& Ours         & \textbf{0.673} & \textbf{1.218} & \textbf{0.309} & \textbf{0.656} \\
\midrule
\multirow{3}{*}{Z-Image}
& Vanilla      & 0.756 & 0.810 & 0.279 & 0.746 \\
& FK-Diffusion & 0.628 & 0.674 & 0.269 & 0.616 \\
& Ours         & \textbf{0.775} & \textbf{0.977} & \textbf{0.285} & \textbf{0.763} \\
\bottomrule
\end{tabular}}
\end{small}
\vskip -0.1in
\end{table}

\begin{table}[t]
\caption{Task-wise accuracy results on the GenEval~\citep{ghosh2023geneval} benchmark.}
\label{tab:task_wise}

  \begin{center}
    \begin{small}
      \resizebox{.85\linewidth}{!}{
        \begin{tabular}{l c c c c c c c}
        \toprule
        &\multicolumn{4}{c}{Objects} & \\
        \cmidrule(lr){3-4}
        Model & Overall & Single & Two  & Counting & Colors & Position & Color Attr. \\
        \midrule
        minDALL-E & 0.23 &0.73 &0.11& 0.12& 0.37& 0.02 &0.01\\
        SD v1.5 & 0.43& 0.97& 0.38 &0.35& 0.76& 0.04& 0.06 \\
        PixArt-alpha &0.48 &0.98& 0.50 &0.44 &0.80& 0.08 &0.07\\
        SD v2.1 &0.50& 0.98& 0.51 &0.44& 0.85& 0.07 &0.17\\
        DALL-E 2 &0.52& 0.94 &0.66& 0.49 &0.77 &0.10 &0.19\\
        SDXL &0.55& 0.98 &0.74& 0.39& 0.85 &0.15& 0.23\\
        SDXL Turbo &0.55& 1.00 &0.72& 0.49& 0.80& 0.10 &0.18\\
        IF-XL &0.61 &0.97 &0.74& 0.66& 0.81 &0.13& 0.35\\
        DALL-E 3 & 0.67& 0.96 &0.87 &0.47& 0.83& 0.43& 0.45\\
        \midrule
        \textbf{SD3} & 0.72&
        \textbf{1.00} & 0.88 & 0.67 & 0.83 & 0.29 & 0.63 \\
        + FK-Diffusion & 0.64&
        0.98 & 0.84 & 0.56 & 0.76 & 0.28 & 0.40 \\
        + Ours & \textbf{0.75} &
        0.99 & \textbf{0.90} & \textbf{0.75} & \textbf{0.85} & \textbf{0.36} & \textbf{0.65} \\
        \midrule
        \textbf{FLUX.1-dev} &0.66&0.99&0.81&0.76&\textbf{0.78}&0.20&\textbf{0.45} \\
        + FK-Diffusion&0.47&0.89&0.56&0.51&0.51&0.15&0.18 \\
        + Ours &\textbf{0.67}&\textbf{0.99}&\textbf{0.86}& \textbf{0.79}&0.78&\textbf{0.21}&0.42 \\
        
        \midrule
        \textbf{Z-Image} 
        &0.76&1.00&0.89&0.75&0.85&\textbf{0.47}&0.57\\
        + FK-Diffusion& 0.63&0.96&0.83&0.66&0.57&0.45&0.31 \\
        + Ours & \textbf{0.78}& \textbf{1.00}&\textbf{0.93}&\textbf{0.86}&\textbf{0.87}&0.42& \textbf{0.57} \\
        \bottomrule
        \end{tabular}}
    \end{small}
  \end{center}
  \vskip -0.1in
\end{table}

\subsection{Experimental Setup}
\label{subsec:exp_setup}

\paragraph{Benchmarks and metrics.}
Following \citet{singhal2025general} and \citet{esser2024scaling}, we evaluate our method on GenEval~\citep{ghosh2023geneval} and T2I-CompBench~\citep{huang2023t2icompbench}. For GenEval, we report the overall and task-wise accuracy together with ImageReward~\citep{xu2023imagereward}, Human Preference Score v2 (HPSv2)~\citep{wu2023human}, and Correct Image rate~\citep{ghosh2023geneval}. For T2I-CompBench, we report BVQA, ImageReward, and HPSv2. All metrics are computed using their official implementations with default settings. ImageReward is used as the reward function for particle weighting and resampling, while all other metrics are used only for evaluation.

\paragraph{Baselines.}
We evaluate our method on three text-to-image flow matching backbones: Stable Diffusion 3 (SD3)\citep{esser2024scaling}, FLUX.1-dev\citep{labs2025flux}, and Z-Image~\citep{team2025zimage}. For each backbone, we compare the original deterministic inference pipeline, FK-Diffusion~\citep{singhal2025general}, and our noise-rejuvenated SMC. We also compare with other text-to-image models, including minDALL-E~\citep{ramesh2022hierarchical}, Stable Diffusion v1.5 (SD v1.5)\citep{rombach2022high}, PixArt-$\alpha$\citep{chen2023pixartalpha}, Stable Diffusion v2.1 (SD v2.1)\citep{rombach2022high}, DALL-E 2\citep{ramesh2022hierarchical}, SDXL~\citep{podell2023sdxl}, SDXL Turbo~\citep{podell2023sdxl}, IF-XL~\citep{saharia2022photorealistic}, and DALL-E 3~\citep{betker2023improving}.

\paragraph{Implementation details.}
For all experiments, the number of inference steps is set to 50. For SMC-based methods, we use $N=4$ particles and apply resampling every 5 steps within the window from step 5 to step 35. For the original deterministic pipeline, we generate 4 independent samples with different noise initializations to match the number of generated candidates used by SMC-based methods. The guidance strength is set to $\lambda=10.0$. Unless otherwise stated, our method uses the look-ahead refinement described in \Cref{subsec:LA}.

\paragraph{Evaluation protocol.}
For each prompt, our method produces $N$ final particles. We decode and evaluate each particle independently, and report the average of each evaluation metric over all particles and prompts. We apply the same protocol to the corresponding vanilla pipeline by generating $N$ independent samples. We do not select samples according to their rewards or aggregate the results using the particle weights.

\begin{table}[t]
\centering
\caption{Comparison on T2I-CompBench~\citep{huang2023t2icompbench} benchmark.}
\label{tab:t2i_bench}
\resizebox{.99\linewidth}{!}{
\begin{tabular}{l ccc ccc ccc}
\toprule
\multirow{2}{*}{Method} 
& \multicolumn{3}{c}{Color} 
& \multicolumn{3}{c}{Shape} 
& \multicolumn{3}{c}{Texture} \\
\cmidrule(lr){2-4} \cmidrule(lr){5-7} \cmidrule(lr){8-10}
& BVQA & ImageReward & HPSv2
& BVQA & ImageReward & HPSv2 
& BVQA & ImageReward & HPSv2 \\
\midrule
SD3          & 0.812 & 1.443 & 0.314 & 0.590 & 0.977 & 0.282 & 0.734 & 1.110 & 0.280 \\
+ Ours   & \textbf{0.820} & \textbf{1.563} & \textbf{0.318} & \textbf{0.641} & \textbf{1.280} & \textbf{0.290} & \textbf{0.760} & \textbf{1.327} & \textbf{0.288} \\
\midrule
FLUX.1-dev        & 0.768 & 1.342 & \textbf{0.312} & 0.497 & 0.691 & 0.274 & 0.635 & 0.856 & 0.286 \\
+ Ours & \textbf{0.915} & \textbf{1.465} & 0.310 & \textbf{0.536} & \textbf{0.975} & \textbf{0.280} & \textbf{0.650} & \textbf{1.164} & \textbf{0.290} \\
\midrule
Z-Image        & 0.821 & 0.925 & 0.291 & 0.606 & 0.665 & 0.256 & 0.761 & 0.580 & 0.254 \\
+ Ours & \textbf{0.823} & \textbf{1.069} & \textbf{0.294} & \textbf{0.634} & \textbf{0.901} & \textbf{0.261} & \textbf{0.787} & \textbf{0.764} & \textbf{0.260} \\
\bottomrule
\end{tabular}}
\end{table}

\subsection{Comparison Results}
\label{subsec:comparison_results}

\paragraph{GenEval evaluation.}
The main quantitative results on GenEval are summarized in Table~\ref{tab:overall}. Across all three flow-matching backbones, our method consistently improves over the original deterministic inference pipeline. On SD3~\citep{esser2024scaling}, noise-rejuvenated SMC improves the GenEval score from $0.718$ to $0.748$ and the ImageReward score from $1.081$ to $1.268$. On FLUX.1-dev~\citep{labs2025flux}, our method improves ImageReward from $0.986$ to $1.218$, while also improving GenEval and HPSv2. On Z-Image~\citep{team2025zimage}, our method improves GenEval from $0.756$ to $0.775$ and ImageReward from $0.810$ to $0.977$. In contrast, FK-Diffusion~\citep{singhal2025general} consistently degrades GenEval and Correct Image across these backbones, suggesting that SMC methods designed for stochastic diffusion samplers do not directly transfer to deterministic flow-matching inference.

We further report the task-wise GenEval results in Table~\ref{tab:task_wise}. Our method achieves improved or comparable accuracy across most categories, with particularly clear gains on two-object generation, counting, and position. These categories require object-level compositional alignment and are sensitive to missing objects or incorrect spatial relations. The improvements on GenEval, HPSv2, and Correct Image are especially important because these metrics are used only for evaluation and not for particle weighting or resampling.

\paragraph{T2I-CompBench evaluation.}
We additionally evaluate compositional generation on T2I-CompBench in Table~\ref{tab:t2i_bench}. Across color, shape, and texture composition, our method improves most metrics on SD3, FLUX.1-dev, and Z-Image. In particular, the improvements on BVQA and HPSv2 provide evidence beyond the ImageReward guidance signal. These results suggest that noise-rejuvenated SMC improves reward-guided generation while also transferring to broader compositional evaluation settings.

\begin{figure}[t]
  \begin{center}
\centerline{\includegraphics[width=.99\linewidth]{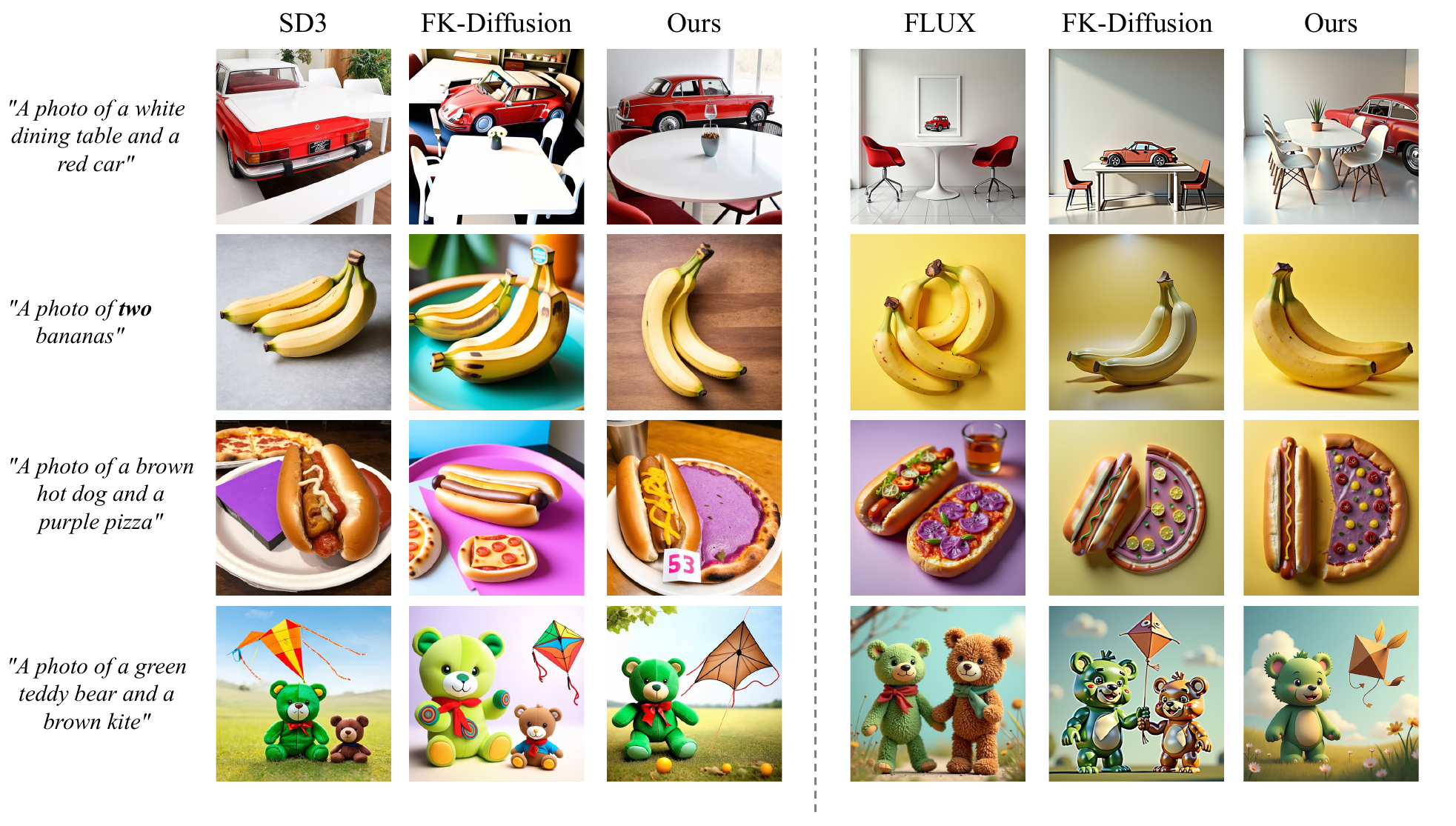}}
    \caption{Qualitative comparison across different inference pipelines. Results are shown for Stable Diffusion 3 (SD3)~\citep{esser2024scaling} and FLUX.1-dev~\citep{labs2025flux} backbones under the same prompts. For each selected prompt, the displayed samples are uncurated outputs from the corresponding inference pipeline. Our method generates samples that more closely follow the prompts in terms of semantic attributes and composition, while preserving visual quality.}
    \label{fig:main_ret}
  \end{center}
  \vskip -0.2in
\end{figure}

\begin{figure}[t]
  \begin{center}
\centerline{\includegraphics[width=.99\linewidth]{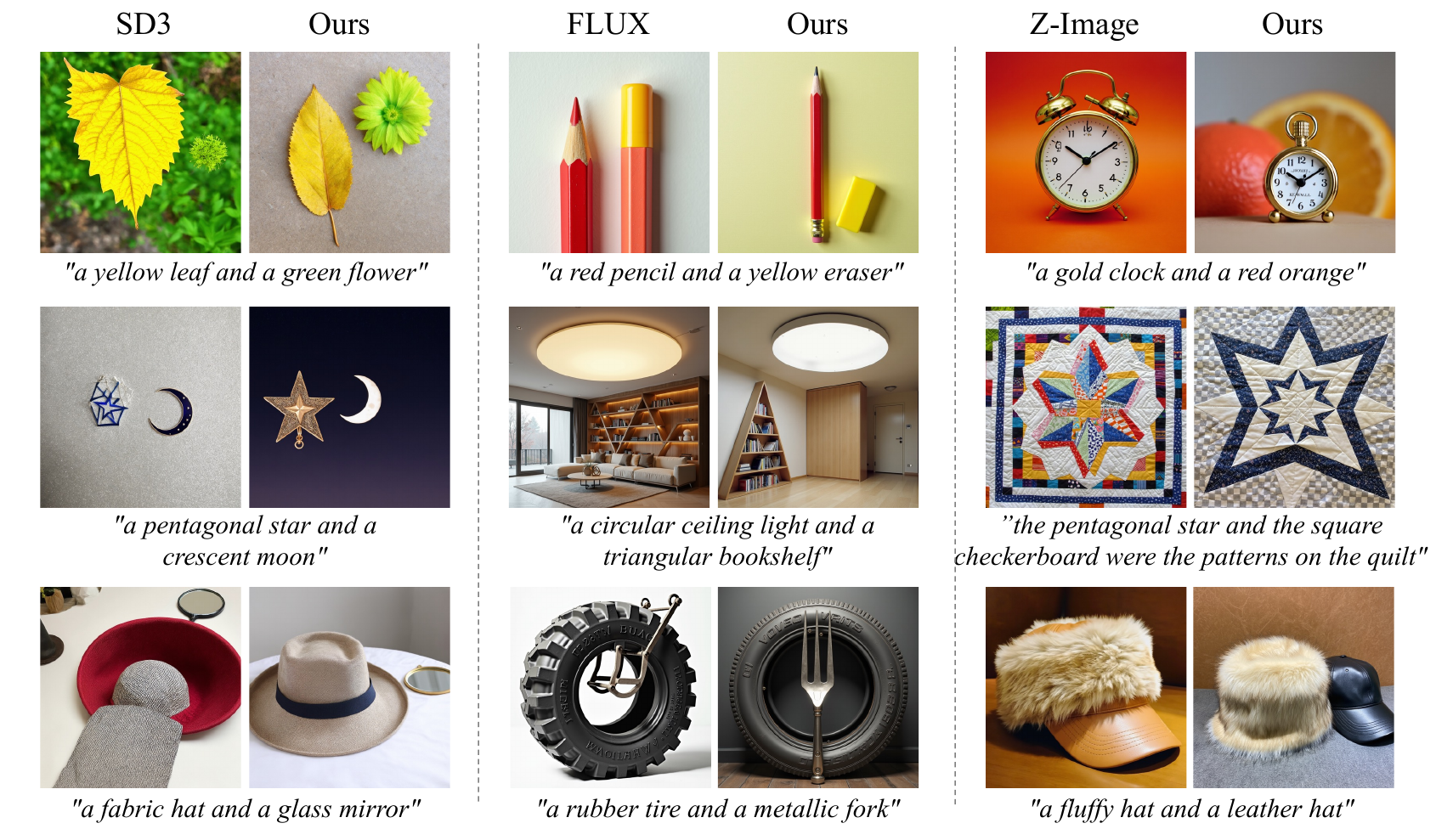}}
    \caption{Qualitative comparison across different inference pipelines on T2I-CompBench. Our method generates samples that more closely follow the prompts in terms of semantic attributes and composition, while preserving visual quality.}
    \label{fig:t2i-bench}
  \end{center}
  \vskip -0.1in
\end{figure}

\paragraph{Qualitative results.}
Qualitative comparisons are shown in Figure~\ref{fig:main_ret} and Figure~\ref{fig:t2i-bench}. We compare samples generated by the original SD3 and FLUX.1-dev pipelines, FK-Diffusion, and our method under the same prompts. FK-Diffusion introduces additional stochasticity but can produce artifacts or reduce image consistency when directly applied to flow-matching samplers. For example, it may produce incorrect object counts or unnatural textures, such as the banana example with SD3 and the teddy-bear example with FLUX. In contrast, our method tends to better follow the prompts in terms of object counts, color attributes, and spatial composition, while preserving visual quality. Additional qualitative examples are provided in Appendix~\ref{appsec:more_rets}.

%% file: sec/5-discussion.tex
\subsection{Discussion and Analysis}

\begin{figure}[t]
  \centering

  \begin{minipage}[t]{0.49\linewidth}
    \vspace{0pt}
    \centering

    \captionof{table}{Effect of Noise Rejuvenation (NR.). Directly applying an SDE solver degrades performance.}
    \label{tab:rejuvenation_check}
    \begin{small}
      \resizebox{0.95\linewidth}{!}{
        \begin{tabular}{lccc}
          \toprule
          Method & GenEval & ImageReward & HPSv2 \\
          \midrule
          SD3 & \textbf{0.718} & \textbf{1.081} & 0.302 \\
          SD3 + SDE solver & 0.585 & 0.799 & 0.286 \\
          SD3 + NR. (Ours) & 0.706 & 1.065 & \textbf{0.304} \\
          \bottomrule
        \end{tabular}}
    \end{small}

    \vspace{0.05in}

    \includegraphics[width=\linewidth]{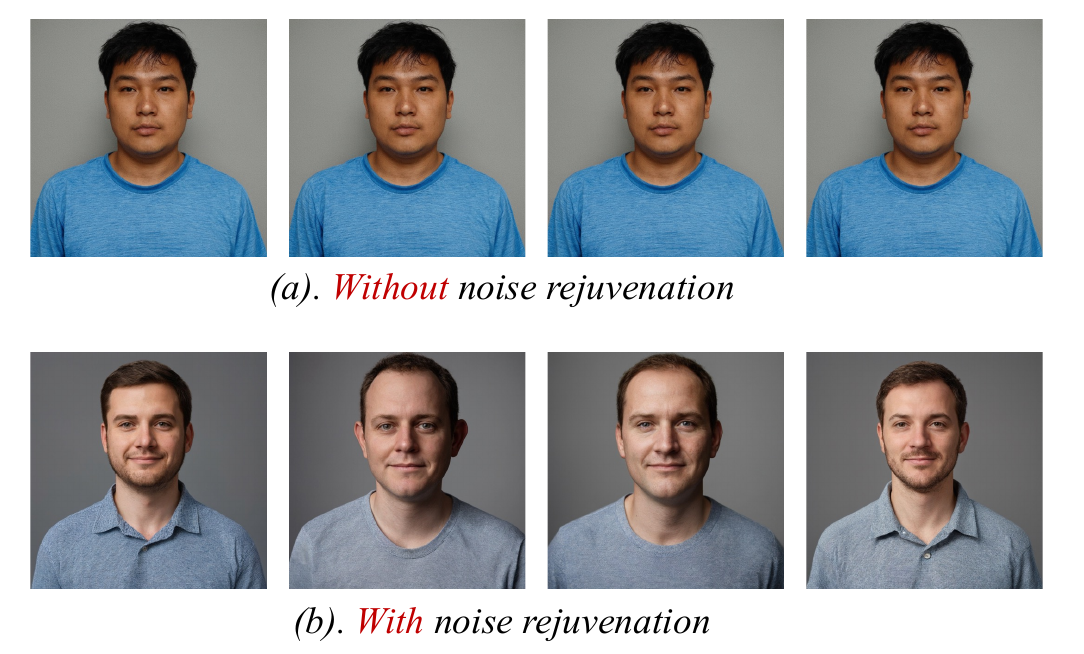}
    \captionof{figure}{An illustration of using SMC without and with noise rejuvenation for flow matching.}
    \label{fig:ode}
  \end{minipage}
  \hfill
  \begin{minipage}[t]{0.48\linewidth}
    \vspace{0pt}
    \centering

    \includegraphics[width=\linewidth]{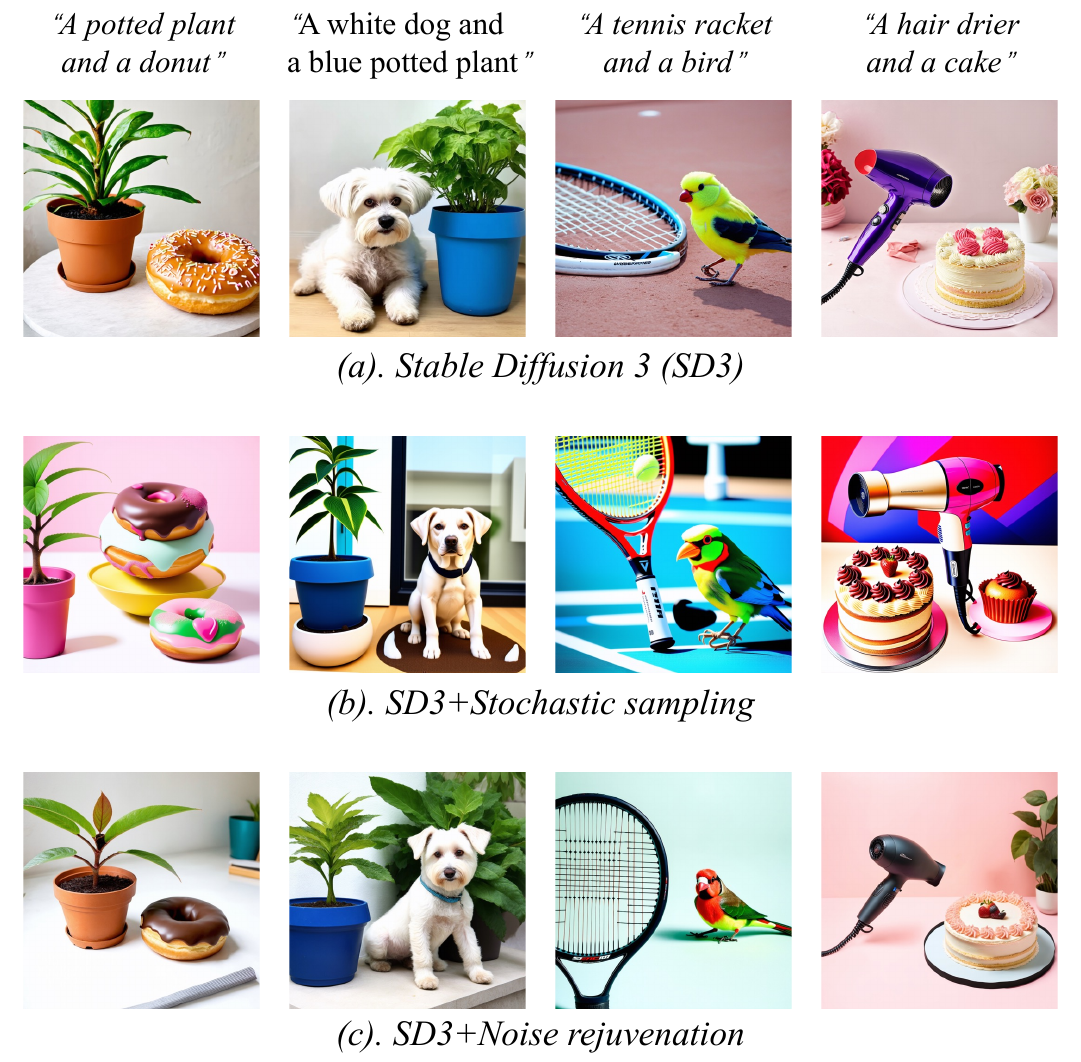}
    \captionof{figure}{Comparison of stochastic sampling and our noise rejuvenation on flow matching.}
    \label{fig:noise}
  \end{minipage}

  \vskip -0.1in
\end{figure}

\paragraph{Effect of noise rejuvenation.}
Directly applying SMC to deterministic flow matching leads to particle collapse after resampling, causing all particles to follow the same trajectory and converge to identical samples, as illustrated in Figure~\ref{fig:ode}. Noise rejuvenation alleviates this issue by introducing stochasticity only during resampling, preserving particle diversity and enabling exploration along distinct trajectories.

To isolate the effect of noise rejuvenation, we apply it to the original SD3 pipeline without SMC. As shown in Table~\ref{tab:rejuvenation_check}, noise rejuvenation only slightly affects generation
performance, suggesting that the introduced
approximation errors are limited and the underlying
flow matching behavior is largely preserved. In contrast, injecting noise at every sampling step through an SDE solver significantly degrades generation quality (Figure~\ref{fig:noise}). These results suggest that, in this setting, stochasticity is beneficial only when introduced selectively, rather than throughout the entire sampling process.

\paragraph{Effect of look-ahead.}
\label{subsec:effect_la}

We evaluate the effect of the look-ahead mechanism in Table~\ref{tab:efficiency}. Under the default setting with $N=4$ particles, enabling look-ahead improves all reported metrics on both SD3 and FLUX.1-dev. This suggests that evaluating rejuvenated particles after a short flow-matching rollout provides more stable weights for resampling. The improvement comes with additional computation, increasing the number of function evaluations (NFE) and the average inference time per prompt. Thus, look-ahead should be viewed as a compute-quality trade-off rather than a no-cost improvement.

We further compare different particle numbers with and without look-ahead in Table~\ref{tab:particle_num_check}. Look-ahead is most helpful when the number of particles is small, where a more accurate estimate of each rejuvenated particle can improve particle selection. As the number of particles increases, the two variants become comparable, suggesting that additional particles can partly compensate for the lack of look-ahead refinement.

\begin{table}[t]
  \caption{Results of our method with and without look-ahead on SD3 and FLUX baseline models. We further report the number of function evaluations (NFE) and the average inference time per prompt for reference. Average inference times were measured on a single NVIDIA A100 GPU.}
  \label{tab:efficiency}
  \begin{center}
    \begin{small}
      \resizebox{.85\linewidth}{!}{
        \begin{tabular}{llccccc}
          \toprule
          Model & Method & NFE $\downarrow$ & Avg. Time (s)$\downarrow$ & GenEval $\uparrow$ & ImageReward $\uparrow$ & HPSv2 $\uparrow$ \\
          \midrule
          \multirow{2}{*}{SD3}
          & Ours(w/o LA) & 200 & 35.10 & 0.742 & 1.261 & 0.308 \\
          & Ours(w/ LA)  & 228 & 46.25 & 0.748 & 1.268 & 0.309 \\
          \midrule
          \multirow{2}{*}{FLUX.1-dev}
          & Ours(w/o LA) & 200 & 95.96  & 0.667 & 1.192 & 0.307 \\
          & Ours(w/ LA)  & 228 & 109.32 & 0.673 & 1.218 & 0.309 \\
          \bottomrule
        \end{tabular}}
    \end{small}
  \end{center}
  \vskip -0.1in
\end{table}

\begin{table}[t]
\caption{Ablation experiment on the number of particles $N$ with and without look-ahead.}
\centering
\label{tab:particle_num_check}
\begin{small}
\resizebox{0.85\linewidth}{!}{
\begin{tabular}{c ccc ccc}
\toprule
& \multicolumn{3}{c}{\textbf{With Look-ahead}} 
& \multicolumn{3}{c}{\textbf{Without Look-ahead}} \\
\cmidrule(lr){2-4} \cmidrule(lr){5-7}
\textbf{Number of Particles} 
& GenEval & ImageReward & HPSv2
& GenEval & ImageReward & HPSv2 \\
\midrule
$N=\phantom{0}2$  & 0.720 & 1.079 & 0.304 & 0.706 & 1.063 & 0.303 \\
$N=\phantom{0}4$  & 0.748 & 1.268 & 0.309 & 0.742 & 1.261 & 0.308 \\
$N=\phantom{0}8$  & 0.761 & 1.322 & 0.311 & 0.740 & 1.323 & 0.309 \\
$N=16$            & 0.766 & 1.347 & 0.311 & 0.763 & 1.348 & 0.311 \\
\bottomrule
\end{tabular}}
\end{small}
\end{table}

\begin{figure}[t]
  \centering

  \begin{minipage}[t]{0.48\linewidth}
    \vspace{0pt}
    \centering
    \includegraphics[width=\linewidth]{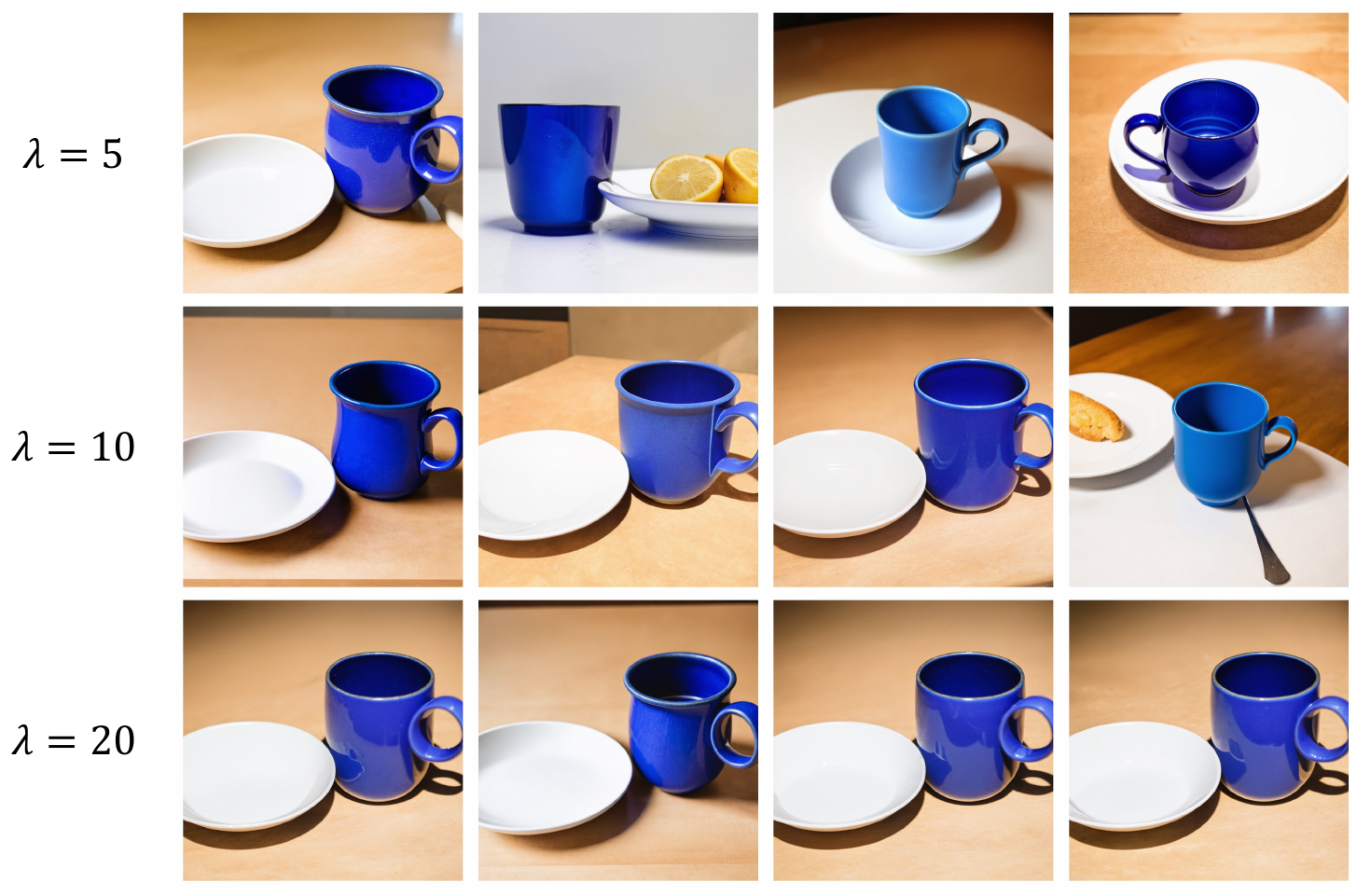}
    \captionof{figure}{Our text-to-image generation results using different reward guidance strengths $\lambda=5,10,20$, with prompt \textit{``a photo of a blue cup next to a white plate''}. Larger $\lambda$ applies stronger reward guidance, leading to samples that more consistently follow the prompt but may slightly reduce generation diversity.}
    \label{fig:lambda}
  \end{minipage}
  \hfill
  \begin{minipage}[t]{0.49\linewidth}
    \vspace{0pt}
    \centering
    \includegraphics[width=\linewidth]{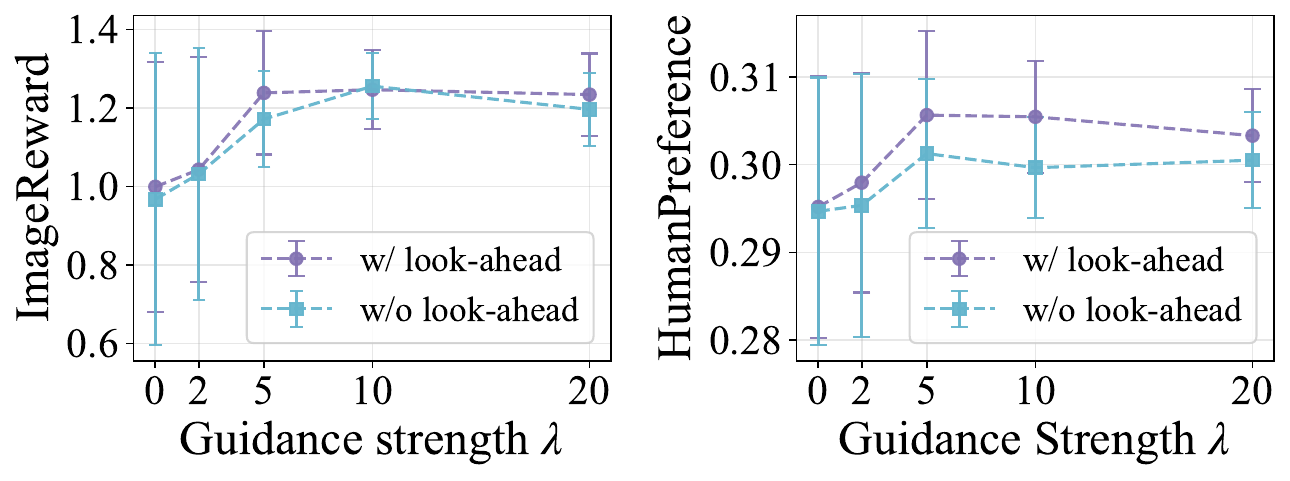}
    \vspace{-0.23in}
    \captionof{figure}{Results of different guidance strengths $\lambda$ for our method with/without look-ahead on SD3.}
    \label{fig:lambda_check}
    \vspace{0.12in}
    \includegraphics[width=\linewidth]{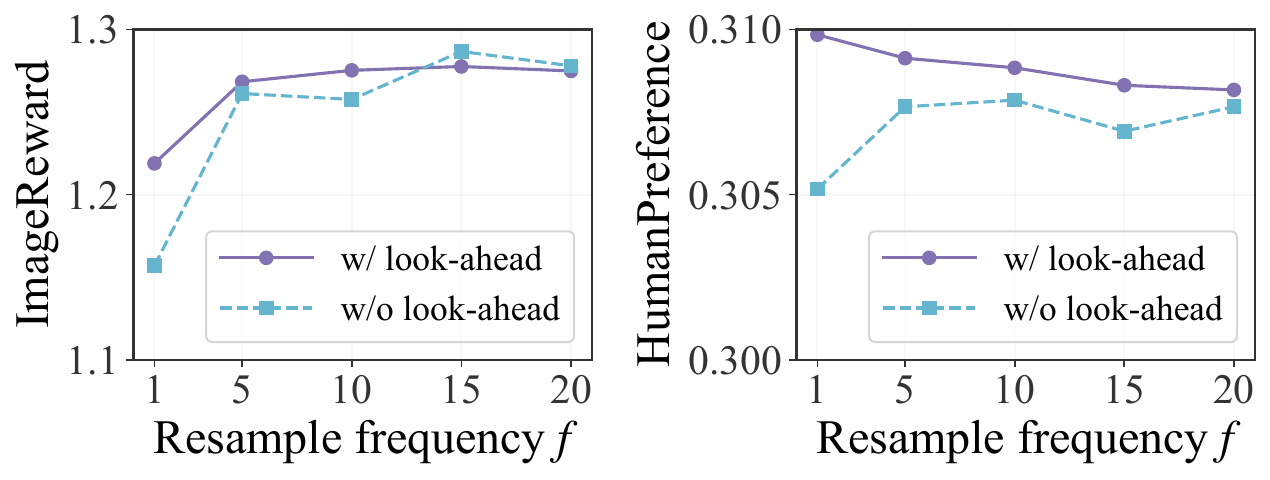}
    \vspace{-0.23in}
    \captionof{figure}{Ablation study on resampling frequency $f$.}
    \label{fig:fre_check}
  \end{minipage}

  \vskip -0.15in
\end{figure}

\paragraph{Effect of guidance strength.}
We examine the effect of the guidance strength $\lambda$ in our framework by varying its value at inference time while keeping all other components fixed. As shown in Figure~\ref{fig:lambda_check}, when $\lambda=0$, corresponding to the absence of guidance, all evaluation metrics attain their lowest values. The performance improves steadily as $\lambda$ increases from 0 to 10, suggesting that setting a suitable guidance strength such as $\lambda=10$ effectively biases the sampling process toward higher-reward regions of the data distribution. We also observe that increasing $\lambda$ to 20 induces higher resampling probability but gives marginal performance gains and reduces sample diversity, as illustrated in Figure~\ref{fig:lambda}. Overall, these results indicate that moderate guidance strengths are sufficient to achieve strong performance in practice.

\paragraph{Effect of resampling frequency.}
We investigate whether the model benefits from more frequent resampling by varying the resampling interval $f$ while keeping the total number of solver steps fixed. As shown in Figure~\ref{fig:fre_check}, the effect of resampling frequency illustrates diminishing returns: ImageReward improves as resampling becomes less frequent and quickly saturates around $f=5$, while Human Preference score shows only minor variation across different frequencies. These results indicate that our method does not require frequent resampling to perform well and remains robust under reduced inference-time budgets.

\paragraph{Effect of particle numbers.}
We study the effect of the number of particles $N$ to characterize inference-time scaling behavior. As shown in \autoref{tab:particle_num_check}, performance consistently improves as $N$ increases from 2 to 16 for both variants, indicating that the proposed noise rejuvenation SMC effectively utilizes additional particles for reward-guided exploration rather than producing redundant samples.

\begin{wraptable}{r}{0.38\linewidth}
  \vspace{-0.05in}
  \caption{Comparison with Best-of-$N$.}
  \vspace{-0.05in}
  \label{tab:best-of-n}
  \centering
  \resizebox{0.99\linewidth}{!}{
  \begin{tabular}{lccc}
    \toprule
    Method  & $N=2$  & $N=4$  & $N=8$  \\
    \midrule
    Best-of-$N$ & 1.227 & 1.369 & 1.438 \\
    Ours        & \textbf{1.255} & \textbf{1.374} & \textbf{1.448} \\
    \bottomrule
  \end{tabular}
  }
  \vspace{-0.1in}
\end{wraptable}

\paragraph{Best-of-N comparison and diversity analysis.}
We also compare with vanilla Best-of-$N$ sampling (i.e., only evaluate the highest-reward sample among $N$ independent samples) in Table~\ref{tab:best-of-n}. Unlike SMC, Best-of-N does not use intermediate resampling or particle interaction during generation. Our method achieves higher best-sample ImageReward than Best-of-$N$ for $N=2,4,8$. This suggests that the improvement is not merely due to drawing multiple independent samples, but comes from particle interaction and rejuvenated exploration during the generation process.

\begin{wraptable}{r}{0.4\textwidth}
\vspace{-.2in}
\centering
\caption{Diversity analysis on GenEval.}
\vspace{-0.05in}
\label{tab:diversity}
\small
\begin{tabular}{lcc}
\toprule
Method & LPIPS$\uparrow$ & IR$\uparrow$ \\
\midrule
\textbf{SD3}              & 0.688 & 1.081 \\
SMC w/o rejuvenation & 0.154 & 1.282 \\
SMC w/ rejuvenation  & 0.367 & 1.268 \\
\midrule
\textbf{FLUX}              & 0.605 & 0.986 \\
SMC w/o rejuvenation & 0.187 & 1.243 \\
SMC w/ rejuvenation  & 0.364 & 1.218 \\
\bottomrule
\end{tabular}
\vspace{-1em}
\end{wraptable}

In addition, we quantify particle diversity using mean pairwise LPIPS among generated samples. As shown in Table~\ref{tab:diversity}, SMC without rejuvenation suffers from severe particle collapse, reducing pairwise LPIPS from $0.688$ to $0.154$ on SD3 and from $0.605$ to $0.187$ on FLUX. Noise rejuvenation substantially restores diversity, increasing pairwise LPIPS to $0.367$ on SD3 and $0.364$ on FLUX. Interestingly, SMC without rejuvenation can still obtain high ImageReward, but it does so by collapsing to a small set of high-reward samples. Noise rejuvenation provides a better balance between reward optimization and particle diversity.

\paragraph{Compute--performance trade-off analysis.}
We further evaluate the practical computation--quality trade-off on SD3 using GenEval and T2I-CompBench. Tables~\ref{tab:sd3_geneval_compute} and~\ref{tab:sd3_t2icomp_compute} report NFE, wall-clock time, peak GPU memory, and quality metrics computed on the ImageReward-selected top-1 output, averaged over three random seeds. HPSv2 is evaluated on the same selected output. We additionally compare with Diffusion Tree Sampling (DTS)~\citep{jain2026diffusion}, a recent tree-based inference-time alignment method that reuses information across multiple generation trajectories. Under the same NFE, both our variant without look-ahead and our NFE-matched variant consistently outperform Best-of-4 and FK-Diffusion across the two benchmarks. The NFE-matched variant has runtime and memory usage comparable to FK-Diffusion and DTS~\citep{jain2026diffusion}, while Best-of-4 remains faster and uses slightly less memory. The full method incurs additional NFEs for look-ahead refinement and achieves the strongest overall performance. These results make explicit the additional computational cost and the corresponding quality improvement.

\begin{table}[t]
\centering
\caption{Compute and performance comparison on SD3 with GenEval. Quality metrics are reported as means over three random seeds. \emph{Ours (NFE-matched)} reduces the number of sampling steps from 50 to 43, resulting in 200 NFEs and matching the NFE of Best-of-4, FK-Diffusion, and the other 200-NFE variants. Wall-clock time and peak GPU memory were measured on a single NVIDIA GH200 GPU.} 
\label{tab:sd3_geneval_compute}

\resizebox{\textwidth}{!}{
\begin{tabular}{lcccccc}
\toprule
Method
& ImageReward $\uparrow$
& HPSv2 $\uparrow$
& NFE $\downarrow$
& Wall-clock time (s) $\downarrow$
& Peak GPU memory (GiB) $\downarrow$\\
\midrule
Vanilla-50 & 1.0773 & 0.3023 & 50 & 3.73 & 20.46 \\
Vanilla-200 & 1.0792 & 0.3032 & 200 & 13.84 & 20.46 \\
Best-of-$4$ & 1.3696 & 0.3097 & 200 & 14.59 & 26.11 \\
FK-Diffusion & 1.2536 & 0.2965 & 200 & 18.53 & 27.78 \\
Diffusion Tree Sampling &1.2344 &0.2951  & 218 &16.40  & 25.16 \\
Ours (w/o Look-ahead) & 1.3734 & 0.3103 & 200 & 18.18 & 27.78 \\
Ours (NFE-matched) & 1.3720 & 0.3102 & 200 & 18.74 & 27.78 \\
Ours & 1.3821 & 0.3117 & 228 & 21.69 & 27.78 \\
\bottomrule
\end{tabular}}
\end{table}

\begin{table}[t]
\centering
\caption{Compute and performance comparison on SD3 with T2I-CompBench. Quality metrics are reported as means over three random seeds. \emph{Ours (NFE-matched)} reduces the number of sampling steps from 50 to 43, resulting in 200 NFEs and matching the NFE of Best-of-4, FK-Diffusion, and the other 200-NFE variants. Wall-clock time and peak GPU memory were measured on a single NVIDIA GH200 GPU.}
\label{tab:sd3_t2icomp_compute}
\resizebox{\textwidth}{!}{
\begin{tabular}{lccccc}
\toprule
Method
& ImageReward $\uparrow$
& HPSv2 $\uparrow$
& NFE $\downarrow$
& Wall-clock time (s) $\downarrow$
& Peak GPU memory (GiB) $\downarrow$\\
\midrule
Vanilla-50 & 1.1841 & 0.2925 & 50 & 3.75 & 20.46 \\
Vanilla-200 & 1.2028 & 0.2937 & 200 & 13.98 & 20.46 \\
Best-of-$4$ & 1.4826  & 0.2979 & 200 & 14.70 & 26.11 \\
FK-Diffusion & 1.2691 & 0.2853 & 200 & 17.82 & 27.78 \\
Diffusion Tree Sampling & 1.2608 & 0.2848 & 218 &  16.43& 25.16 \\
Ours (w/o Look-ahead) & 1.4944 & 0.2995 & 200 & 17.54 & 27.78 \\
Ours (NFE-matched) & 1.4907 & 0.2994 & 200 & 17.57 & 27.78 \\
Ours & 1.5011 & 0.3005 & 228 & 19.72 & 27.78 \\
\bottomrule
\end{tabular}}
\end{table}

%% file: sec/6-conclusion.tex
\section{Conclusion}

We propose noise rejuvenation, a resampling-time perturbation strategy for applying SMC to deterministic flow-matching text-to-image samplers. By replacing the source noise of a resampled particle using its fixed pre-rejuvenation endpoint estimate, our method restores particle exploration without switching the entire sampler to an SDE. We further introduce a look-ahead refinement mechanism for more informative particle selection. Experiments on SD3, FLUX.1-dev, and Z-Image show consistent improvements over vanilla sampling and FK-Diffusion across GenEval, T2I-CompBench, and human-aligned reward metrics.

\paragraph{Limitations and future work.}
Our method requires additional inference cost that grows approximately linearly with the number of particles. Moreover, noise rejuvenation relies on an approximate endpoint estimate and therefore does not guarantee exact preservation of the original terminal distribution. Strong reward guidance may also introduce local artifacts for challenging compositional prompts. Future work will explore more efficient resampling strategies, adaptive reward schedules, and invariant rejuvenation mechanisms that are more tightly coupled with deterministic flow-matching samplers.

\subsubsection*{Broader Impact Statement}
The proposed framework enables reward-guided test-time scaling for flow-matching text-to-image models without requiring model fine-tuning. Its compatibility with black-box reward functions makes it applicable to a broad range of reward-guided generation tasks. The main limitation is the increased inference cost due to multiple particles and look-ahead refinement.

As particle weights and resampling decisions are determined by an external reward model, the generated samples may inherit its biases. Strong reward guidance can further amplify these biases, reduce sample diversity, and concentrate particles on a small set of high-reward outputs. The choice of reward model and guidance strength should therefore be made with care.

\subsubsection*{Acknowledgments}
This research was supported by \emph{Kjell och M{\"a}rta Beijer Foundation}. Computational resources were provided by the National Academic Infrastructure for Supercomputing in Sweden (NAISS), funded by the Swedish Research Council.

%% file: sec/X-appendix.tex
\newpage
\appendix
\onecolumn

\section{Endpoint Estimation in Flow Matching}
\label{appsec:target_estimation}

Given an intermediate state $z_t$ and a marginal velocity field $v^\ast$, the corresponding ODE endpoint is obtained by integrating the velocity field along the remaining trajectory:
\begin{equation}
    z_1^{\mathrm{ODE}}
    =
    z_t+\int_t^1 v^\ast(z_s,s,c)\,ds.
    \label{eq:int_v_star}
\end{equation}
Recall that the conditional flow-matching path in \Cref{eq:fm-t2i} is defined by the linear interpolation~\cite{liu2022flow}:
\begin{equation}
    z_t=tz_1+(1-t)z_0.
    \label{app-eq:linear_interpolation}
\end{equation}
For a particular pair $(z_0,z_1)$, differentiating this interpolation gives the conditional velocity
\begin{equation}
    u_t(z_t\mid z_0,z_1) = \frac{dz_t}{dt} = z_1 - z_0.
\end{equation}
Importantly, this pairwise conditional velocity is different from the marginal velocity field used during inference. Under the standard flow-matching objective, the optimal marginal velocity is
\begin{equation}
    v^\ast(z_t,t,c) = \mathbb{E} \left[z_1-z_0\mid z_t,c\right].
\end{equation}
Note that \Eqref{app-eq:linear_interpolation} also gives $z_1=z_t+(1-t)(z_1-z_0)$. Taking the conditional expectation of this identity given $(z_t,c)$ yields
\begin{equation}
    z_t+(1-t)v^\ast(z_t,t,c)=\mathbb{E}\left[z_1\mid z_t,c \right].
\end{equation}
Replacing $v^\ast$ with the learned velocity field $v_\theta$ therefore gives the endpoint estimator
\begin{equation}
    \hat{z}_1 = z_t+(1-t)v_\theta(z_t,t,c),
\end{equation}
which is \Cref{eq:z1_estimation} in the main paper. In general, $\hat{z}_1$ should be interpreted as an endpoint estimate rather than an exact sample-wise reconstruction. Its error arises from both the conditional uncertainty of $z_1$ given $(z_t,c)$ and the approximation error of the learned velocity field. Exact reconstruction requires stronger conditions, such as a uniquely determined conditional endpoint and an exact learned velocity field. Therefore, holding $\hat z_1$ fixed during the local rejuvenation update does not by itself establish equality between the rejuvenated and original terminal marginals, consistent with the formulation in~\Cref{sec:method}.

\section{Ablation Study: Resampling at Different Stages}
\label{appsec:sampling_window}
We investigate how the timing of resampling along the flow matching trajectory affects performance. Specifically, we fix the length of the resampling window (i.e., $t_{\text{end}} - t_{\text{start}}$) and vary its position to cover early, middle, and late stages of the trajectory.
Results in \autoref{tab:windows_check} show that performance is relatively robust to these choices, while resampling at very late stages tends to be less effective, with mid-range windows achieving slightly better results. One possible explanation is that, at very late stages, particles are already close to high-reward regions and exploration is largely exhausted. As a result, injecting noise at this stage may mainly introduce additional perturbations without promoting further exploration.

\begin{table}[ht]
  \caption{Results under different resampling stages in our noise rejuvenation inference framework. We vary the resampling time window (early, mid, late, uniform, and full) to analyze how resampling at different stages affects generation quality.}
  \label{tab:windows_check}
  \begin{center}
    \begin{small}
      \resizebox{.8\linewidth}{!}{
        \begin{tabular}{l ccc|ccc}
          \toprule
\textbf{Schedule} 
& $f$ & $t_{\text{start}}$ & $t_{\text{end}}$
&GenEval $\uparrow$ & ImageReward $\uparrow$ & HPSv2  $\uparrow$ \\
\midrule

Early        & 5 & 0  & 30 & 0.737& 1.262 $\pm$ 0.108 & 0.309 $\pm$  0.007 \\
Mid (default)& 5 & 5  & 35 & 0.748 & 1.268 $\pm$ 0.106 & 0.309 $\pm$ 0.007 \\
Late         & 5 & 20 & 50 & 0.721 &1.261 $\pm$ 0.123 &  0.305 $\pm$ 0.009 \\
Uniform window  & 5 & 10 & 40 & 0.740 &1.288 $\pm$ 0.102 & 0.308 $\pm$ 0.007 \\
Full window& 5 & 0 &50 & 0.737 & 1.265 $\pm$ 0.113 & 0.309 $\pm$ 0.007\\
\midrule
Narrow, dense & 3 & 10 & 28 & 0.748 & 1.276 $\pm$ 0.106 &  0.308 $\pm$ 0.008\\
Wide, sparse  & 6 & 0  & 36 & 0.744 &1.269 $\pm$ 0.110 & 0.310 $\pm$ 0.007 \\

          \bottomrule
        \end{tabular}}
    \end{small}
  \end{center}
  \vskip -0.1in
\end{table}

\section{Computational Cost with Increasing Particle Numbers}
\label{appsec:cost_particle}
As shown in~\autoref{tab:cost_particle}, increasing the number of particles leads to a near-linear growth in both NFE and inference time. In return, we observe clear improvements in ImageReward, while the gains on GenEval and HPSv2 are more modest and tend to saturate as $N$ increases from 8 to 16.
\begin{table}[ht]
  \caption{Computational cost with increasing particle numbers on SD3.
We report the number of function evaluations (NFE), average inference time per prompt, and generation performance. Average inference times were measured on a single NVIDIA A100 GPU.}
  \label{tab:cost_particle}
  \begin{center}
    \begin{small}
      \resizebox{.9\linewidth}{!}{
        \begin{tabular}{cccccc}
          \toprule
          Number of Particles  & NFE       & Avg. Time (s) & GenEval $\uparrow$ & ImageReward $\uparrow$ & HPSv2 $\uparrow$   \\
         
            \midrule
                $N=\phantom{0}2$ &114& \phantom{0}18.11 & 0.720 & 1.079& 0.304  \\
                $N=\phantom{0}4$  &228& \phantom{0}46.25& 0.748 & 1.268& 0.309  \\
                $N=\phantom{0}8$ &456& \phantom{0}68.14& 0.761& 1.322& 0.311 \\
                $N=16$ &912& 135.59 & 0.766 & 1.347& 0.311  \\            
          \bottomrule
        \end{tabular}}
    \end{small}
  \end{center}
  \vskip -0.1in
\end{table}


\begin{table}[ht]
  \caption{Comparison of initial-noise rejuvenation with direct local perturbation in the SD3 inference pipeline.}
  \label{tab:rejuvenation_check2}
  \begin{center}
    \begin{small}
      \resizebox{.65\linewidth}{!}{
        \begin{tabular}{lccc}
          \toprule
          Method  &  GenEval & ImageReward   & HPSv2  \\
          \midrule
          SD3  &  0.718& 1.081 & 0.302  \\
           \midrule
          Local perturb ($\sigma_t$ = 0.1) & 0.718 & 1.081 & 0.302  \\
          Local perturb ($\sigma_t$ = 0.5) & 0.729 & 1.267 & 0.310   \\
          Local perturb ($\sigma_t$ = 1.0) & 0.589 & 0.794& 0.279   \\
          \midrule
          Noise rejuvenation (Ours) & 0.748&1.268& 0.309 \\
          
          \bottomrule
        \end{tabular}}
    \end{small}
  \end{center}
  \vskip -0.1in
\end{table}
\section{Analysis of Local-Perturbation Rejuvenation}
\label{appsec:alternative}
To better understand the role of the rejuvenation mechanism, we include a diagnostic comparison with directly perturbing the intermediate particle states in the flow matching pipeline. Specifically, let $\{z_t^{i}\}_{i=1}^K$ denote the resampled particles at time $t$. The local perturbation strategy updates each particle independently as
\begin{equation}
    \tilde z_t^i=z_t^i+\sigma_t\epsilon^i, \qquad \epsilon^i\sim\mathcal{N}(0,I).
\end{equation}
where $\sigma_t$ controls the noise magnitude. The look-ahead step mentioned in Section~\ref{subsec:LA} is then applied to propagate the perturbed particles forward along the deterministic flow matching dynamics. In other words, we add noise to~\Cref{eq:fm-t2i} to construct a corresponding SDE of the form
\begin{equation}
    z_{t+1} = z_t + \int_t^{t+1} v_\theta(z_t, t, c) dt \; + \; \sigma_t dW_t,
\end{equation}
where $W$ is a Wiener process. 

While this creates a dynamic trajectory, its effectiveness is highly sensitive to the choice of $\sigma_t$ and the resampling stage. When $\sigma_t$ is small, the perturbations are quickly damped by the deterministic flow dynamics, leading to rapid particle collapse. Conversely, large perturbations may significantly degrade sample quality, even with the look-ahead refinement. The performance of different $\sigma_t$ used in this strategy is summarized in~\autoref{tab:rejuvenation_check2}. These observations motivate our initial-noise rejuvenation strategy, which perturbs the initial noise and propagates the resulting variation to the intermediate state $z_t$ through the linear interpolation structure of flow matching. As shown in~\autoref{tab:rejuvenation_check2}, our strategy provides a better overall trade-off across GenEval, ImageReward, and HPSv2, supporting its use as the default rejuvenation mechanism in our framework.

\FloatBarrier
\section{Additional Qualitative Results}
\label{appsec:more_rets}

\begin{figure}[t]
\centering
\begin{subfigure}{1.0\linewidth}
    \centering
    \setlength{\abovecaptionskip}{0.02in}
    \includegraphics[width=1.\linewidth]{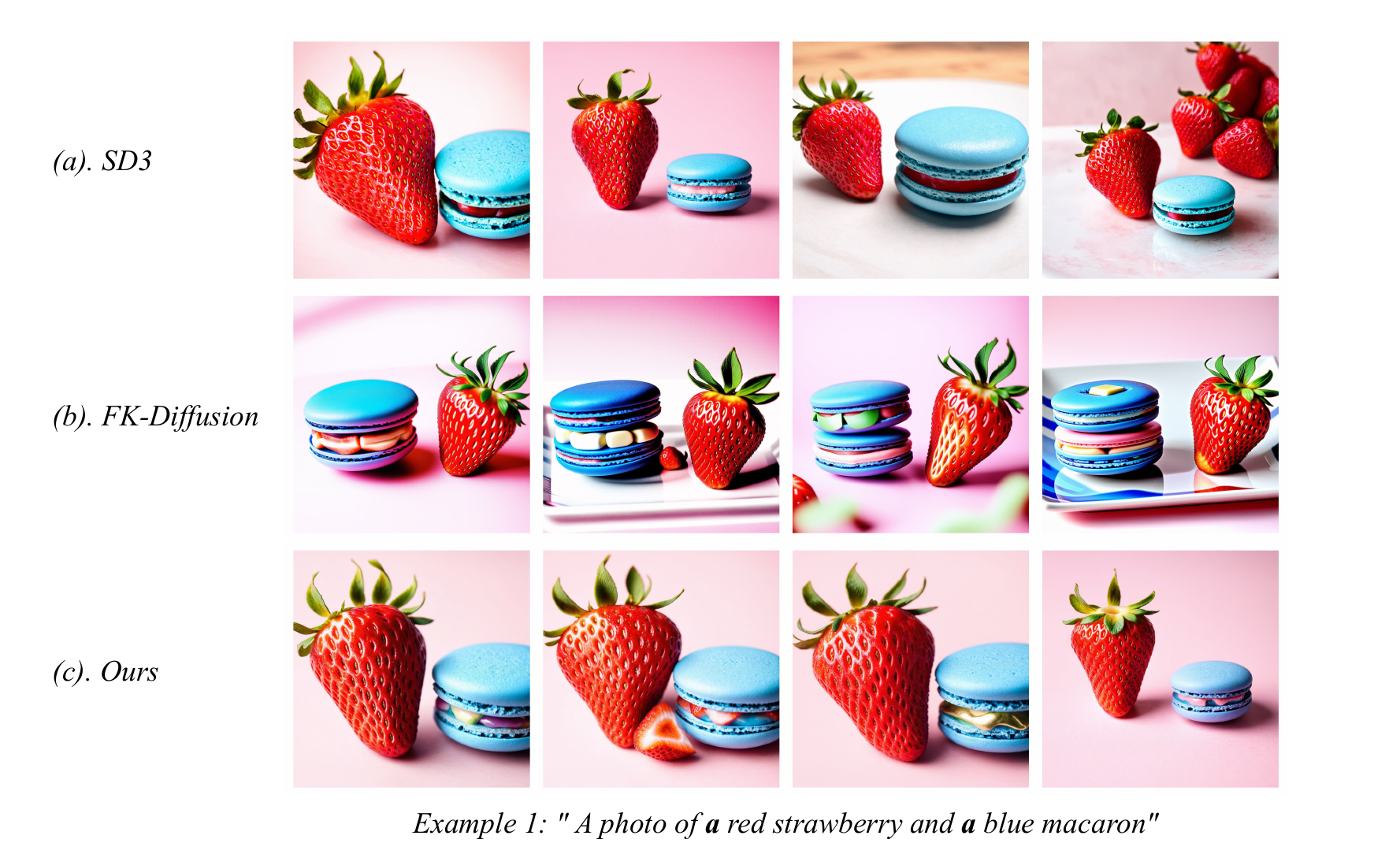}
\end{subfigure}
\begin{subfigure}{1.0\linewidth}
    \centering
    \setlength{\abovecaptionskip}{0.02in}
    \includegraphics[width=1.\linewidth]{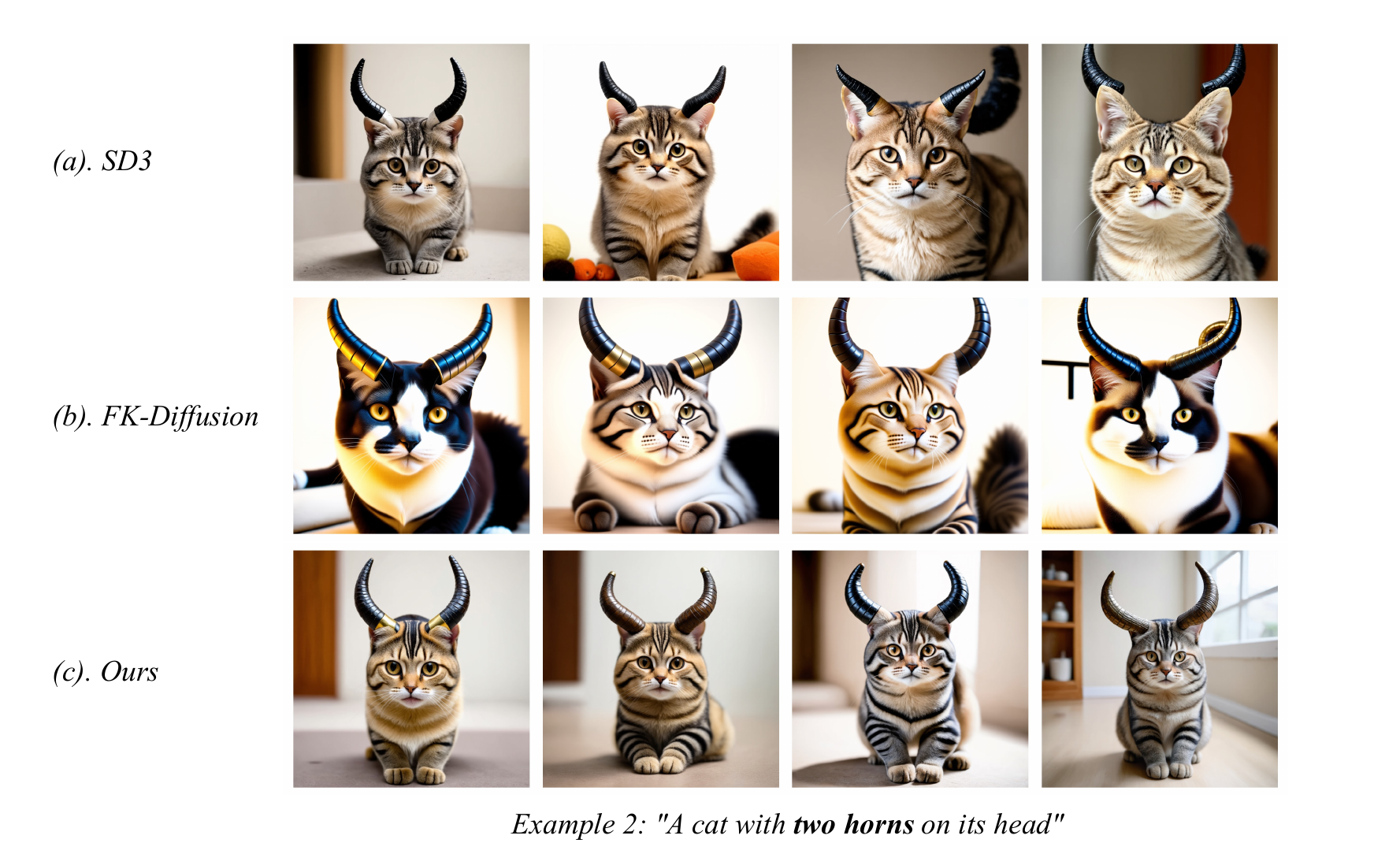}
\end{subfigure}
\caption{\textbf{Additional qualitative results} comparing the vanilla SD3 inference pipeline, FK-Diffusion, and our method on SD3. Please zoom in for details.}
\label{supp-fig:more1}
\end{figure}

\begin{figure}[t]
\centering
\begin{subfigure}{1.0\linewidth}
    \centering
    \setlength{\abovecaptionskip}{0.02in}
    \includegraphics[width=1.\linewidth]{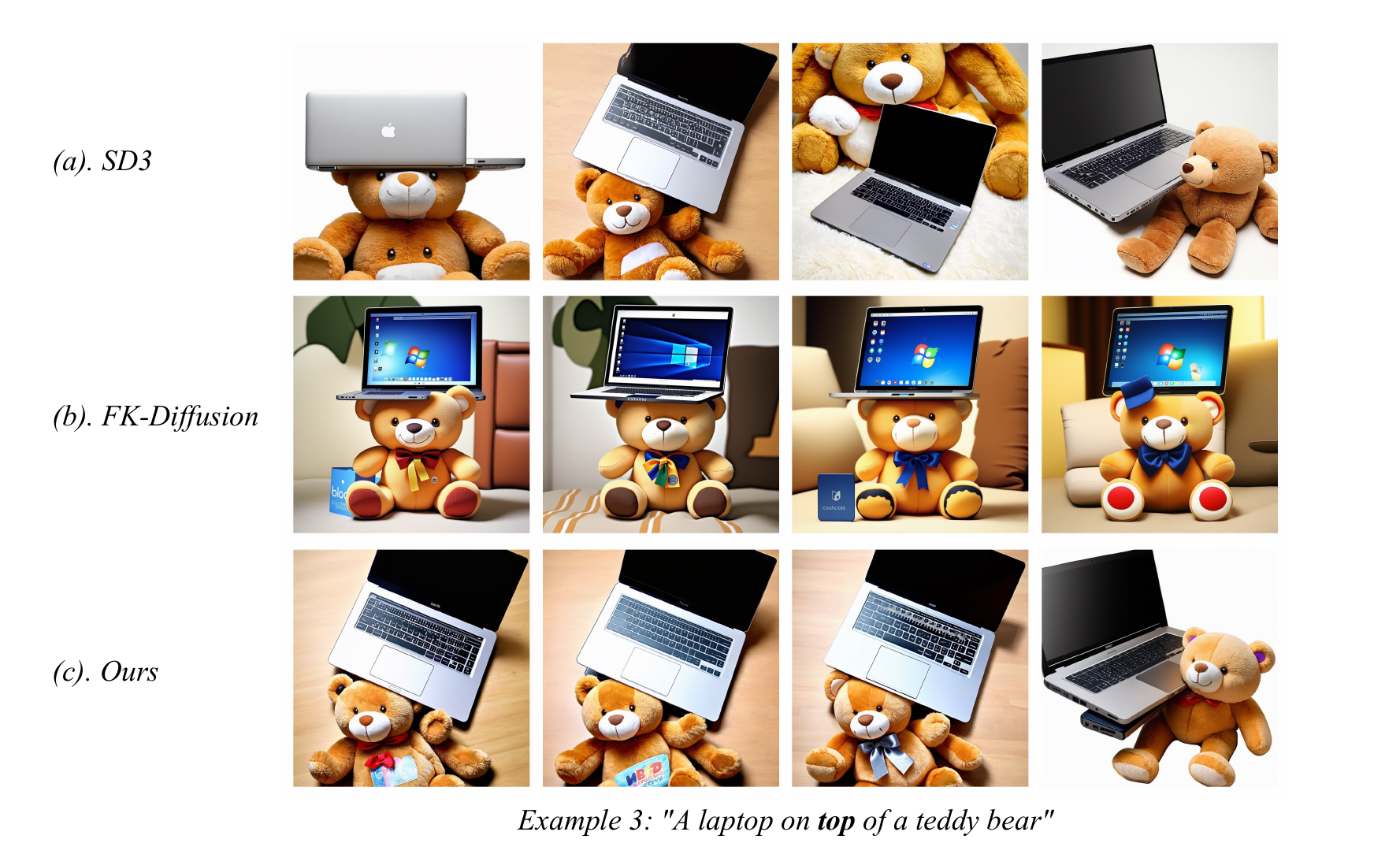}
\end{subfigure}
\begin{subfigure}{1.0\linewidth}
    \centering
    \setlength{\abovecaptionskip}{0.02in}
    \includegraphics[width=1.\linewidth]{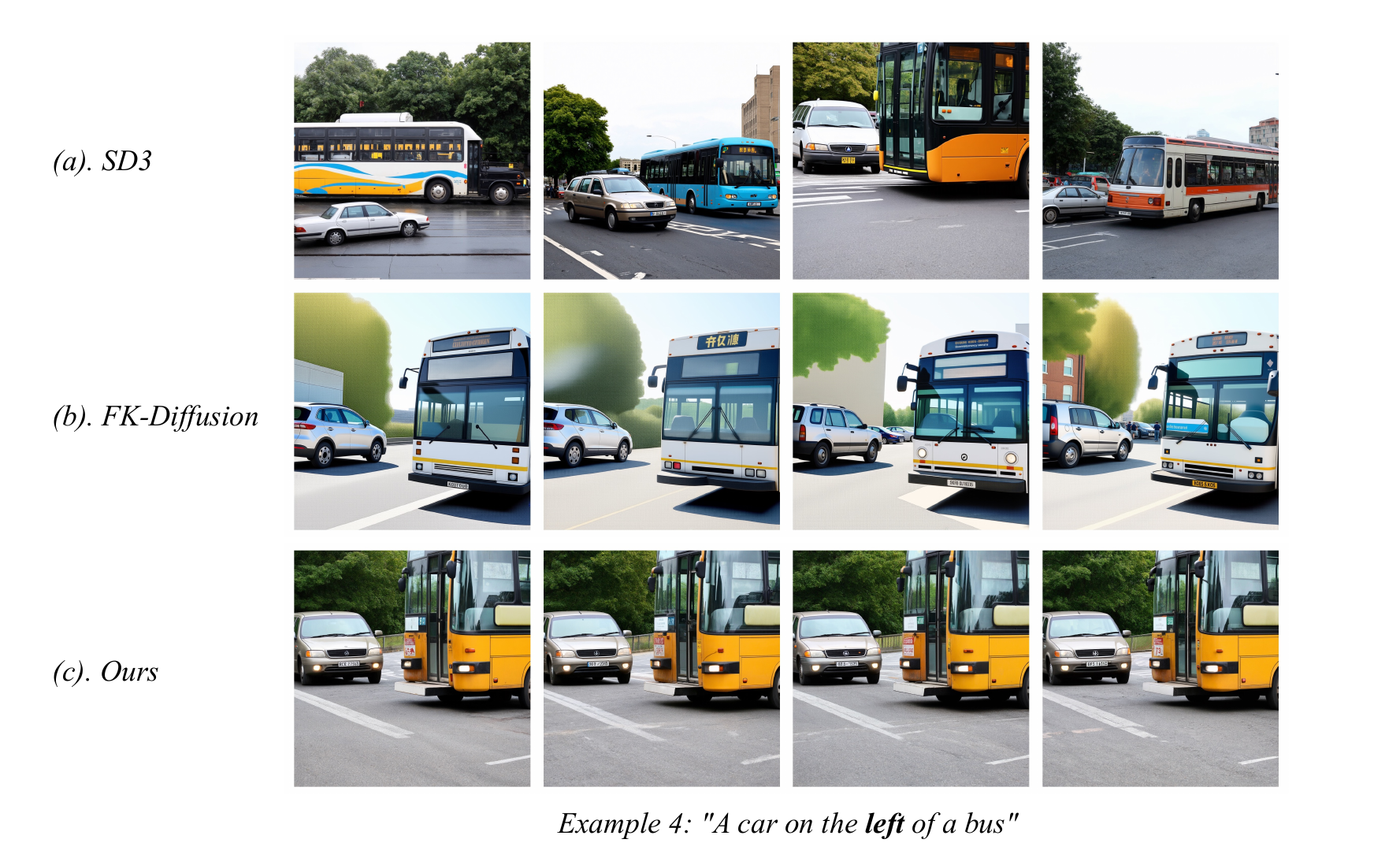}
\end{subfigure}
\caption{\textbf{Additional qualitative results} comparing the vanilla SD3 inference pipeline, FK-Diffusion, and our method on SD3. Please zoom in for details.}
\label{supp-fig:more2}
\end{figure}

We present additional visual comparisons between the vanilla SD3 pipeline, FK-Diffusion, and our method in \autoref{supp-fig:more1} and \autoref{supp-fig:more2}. Here, four particles are used for the SMC-based methods (i.e., FK-Diffusion and ours), while for the vanilla pipeline, we present four independent samples generated by SD3. Prompts are randomly collected from online sources.

\section{Additional Failure Cases and Limitations}
\label{appsec:limitation}
\begin{figure}[t]
\centering
\begin{subfigure}{1.0\linewidth}
    \centering
    \setlength{\abovecaptionskip}{0.02in}
    \includegraphics[width=1.\linewidth]{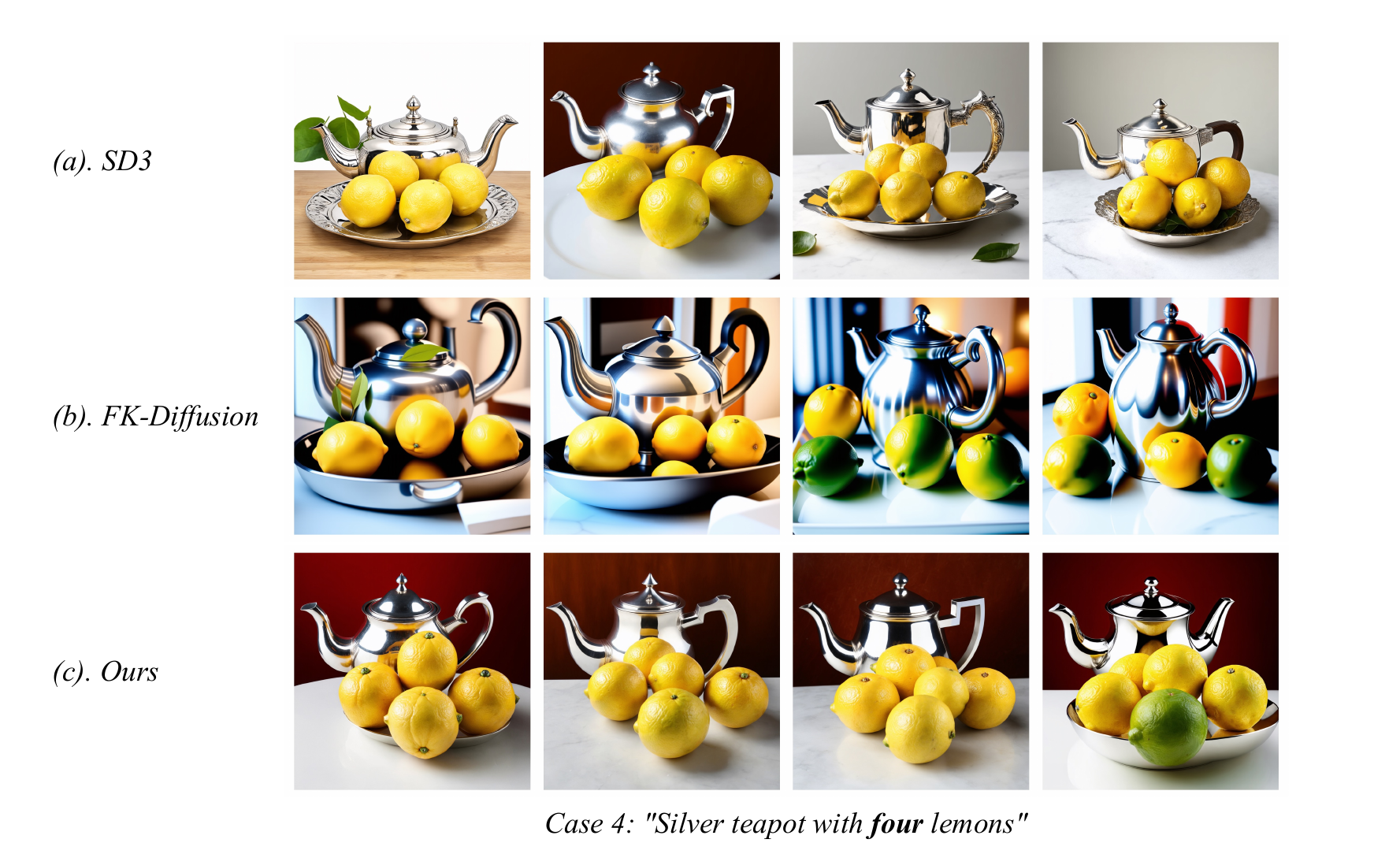}
\end{subfigure}
\begin{subfigure}{1.0\linewidth}
    \centering
    \setlength{\abovecaptionskip}{0.02in}
    \includegraphics[width=1.\linewidth]{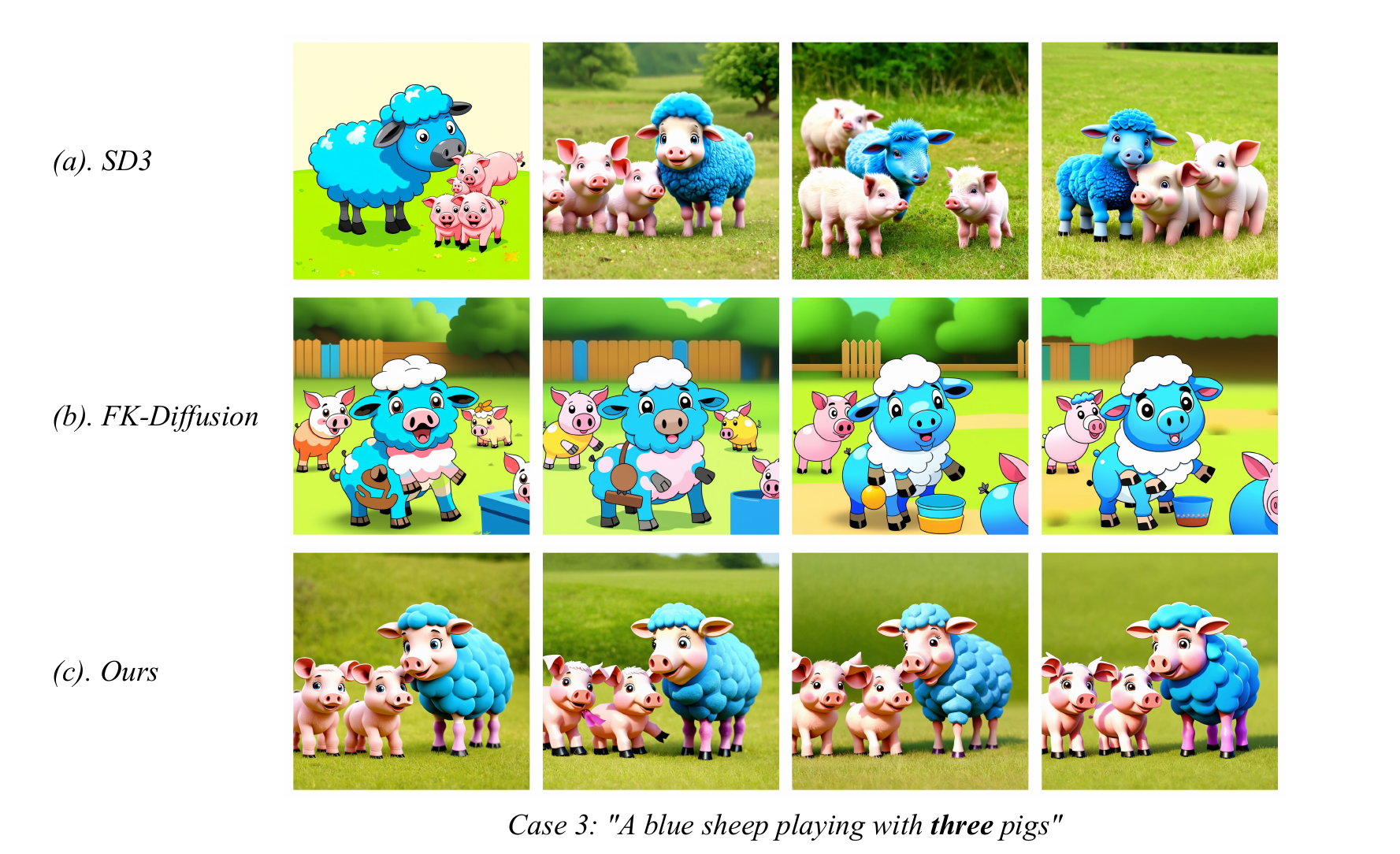}
\end{subfigure}
\caption{\textbf{Challenging cases} for our method on SD3.}
\label{supp-fig:fail1}
\end{figure}

\begin{figure}[t]
\centering
\begin{subfigure}{1.0\linewidth}
    \centering
    \setlength{\abovecaptionskip}{0.02in}
    \includegraphics[width=1.\linewidth]{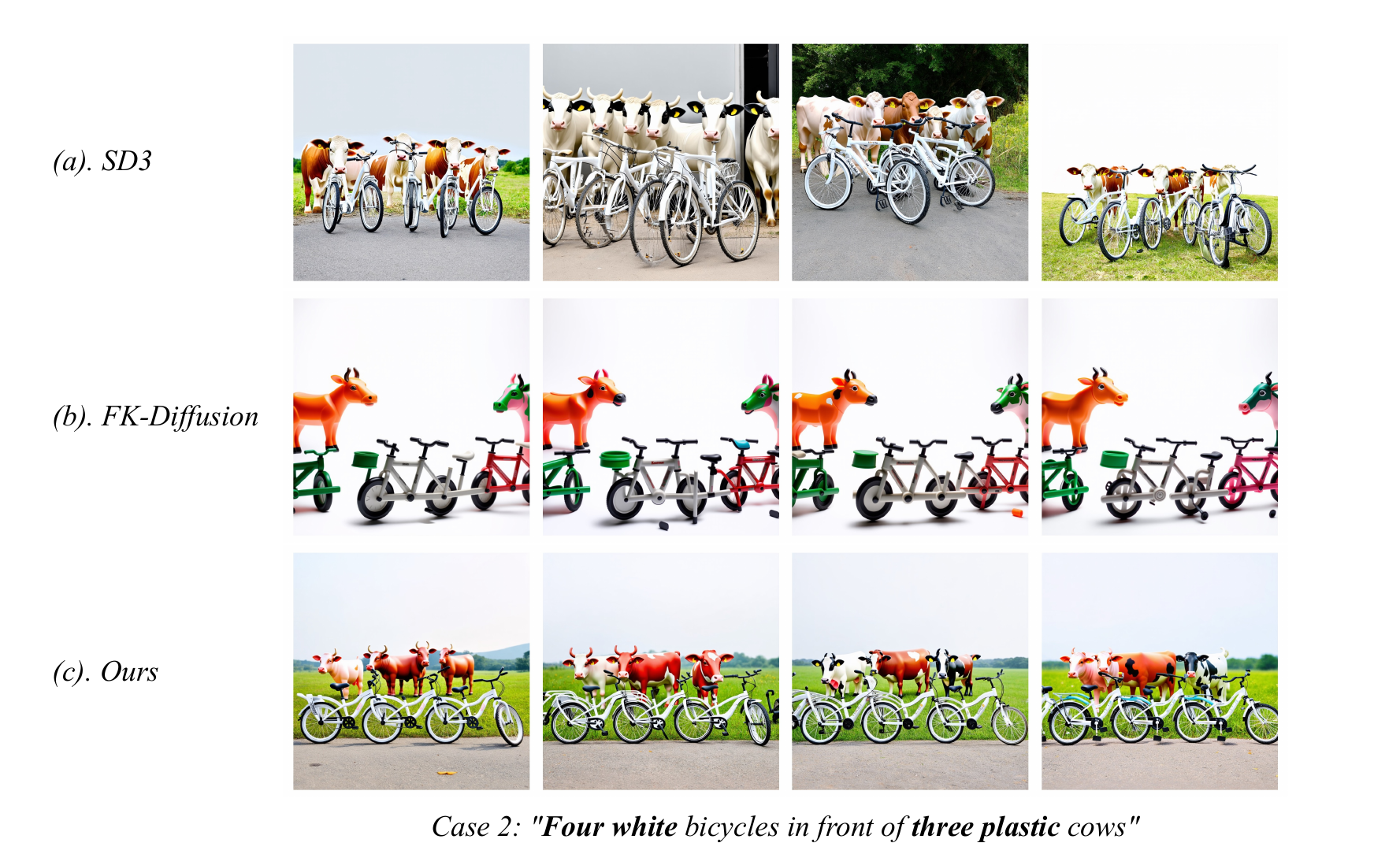}
\end{subfigure}
\begin{subfigure}{1.0\linewidth}
    \centering
    \setlength{\abovecaptionskip}{0.02in}
    \includegraphics[width=1.\linewidth]{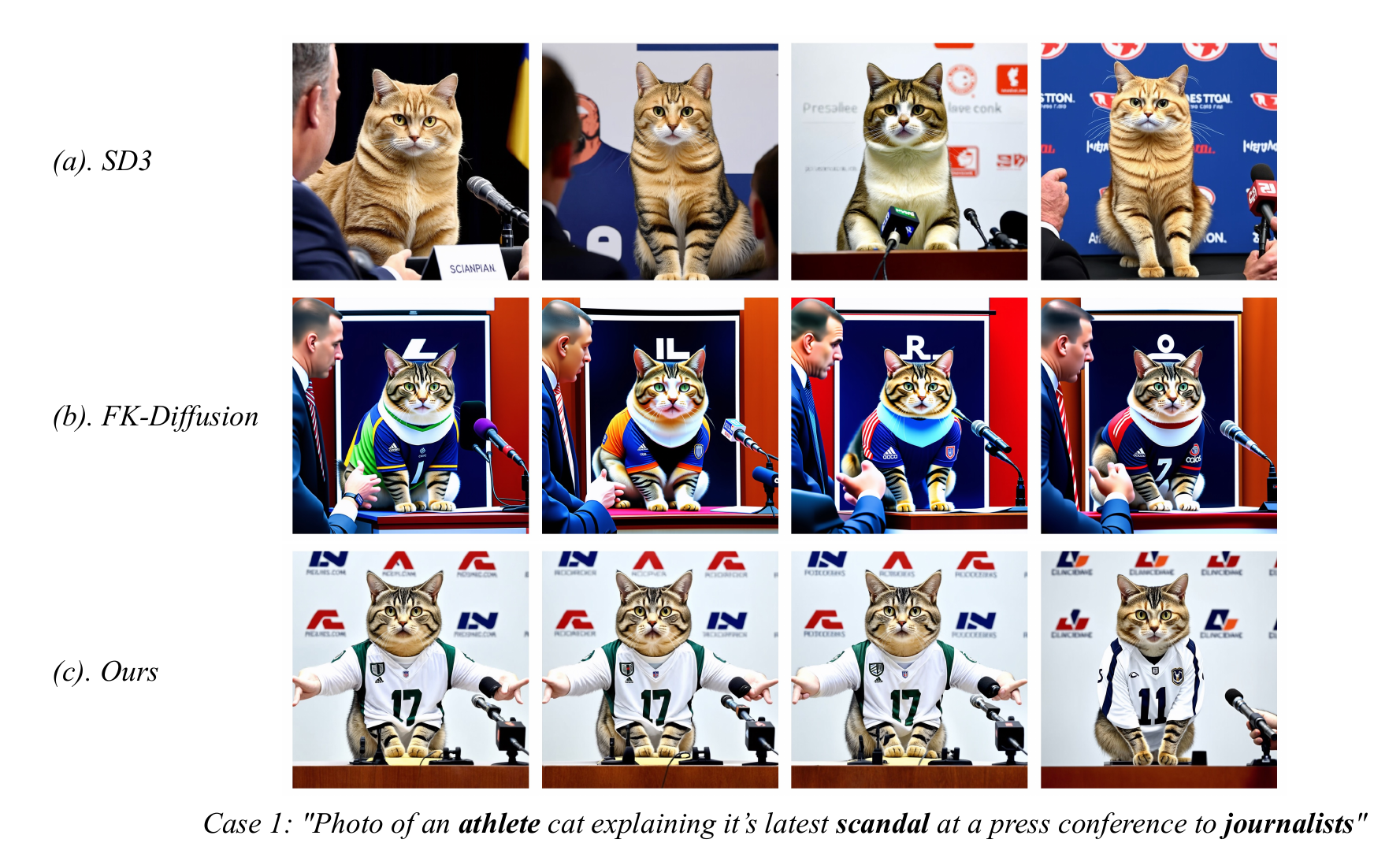}
\end{subfigure}
\caption{\textbf{Challenging cases} for our method on SD3.}
\label{supp-fig:fail2}
\end{figure}

We provide representative failure cases of our method in \autoref{supp-fig:fail1} and \autoref{supp-fig:fail2}. For the SMC-based methods (FK-Diffusion and ours), we use four particles, while for the vanilla pipeline we present four independent SD3 samples. These failure cases mainly arise from complex compositional prompts or prompts describing uncommon scenarios.

\section{Additional Qualitative Comparison of Noise Rejuvenation}
\label{appsec:NR}
\begin{figure}[t]
\centering
\begin{subfigure}{0.85\linewidth}
    \centering
    \setlength{\abovecaptionskip}{0.02in}
    \includegraphics[width=1.\linewidth]{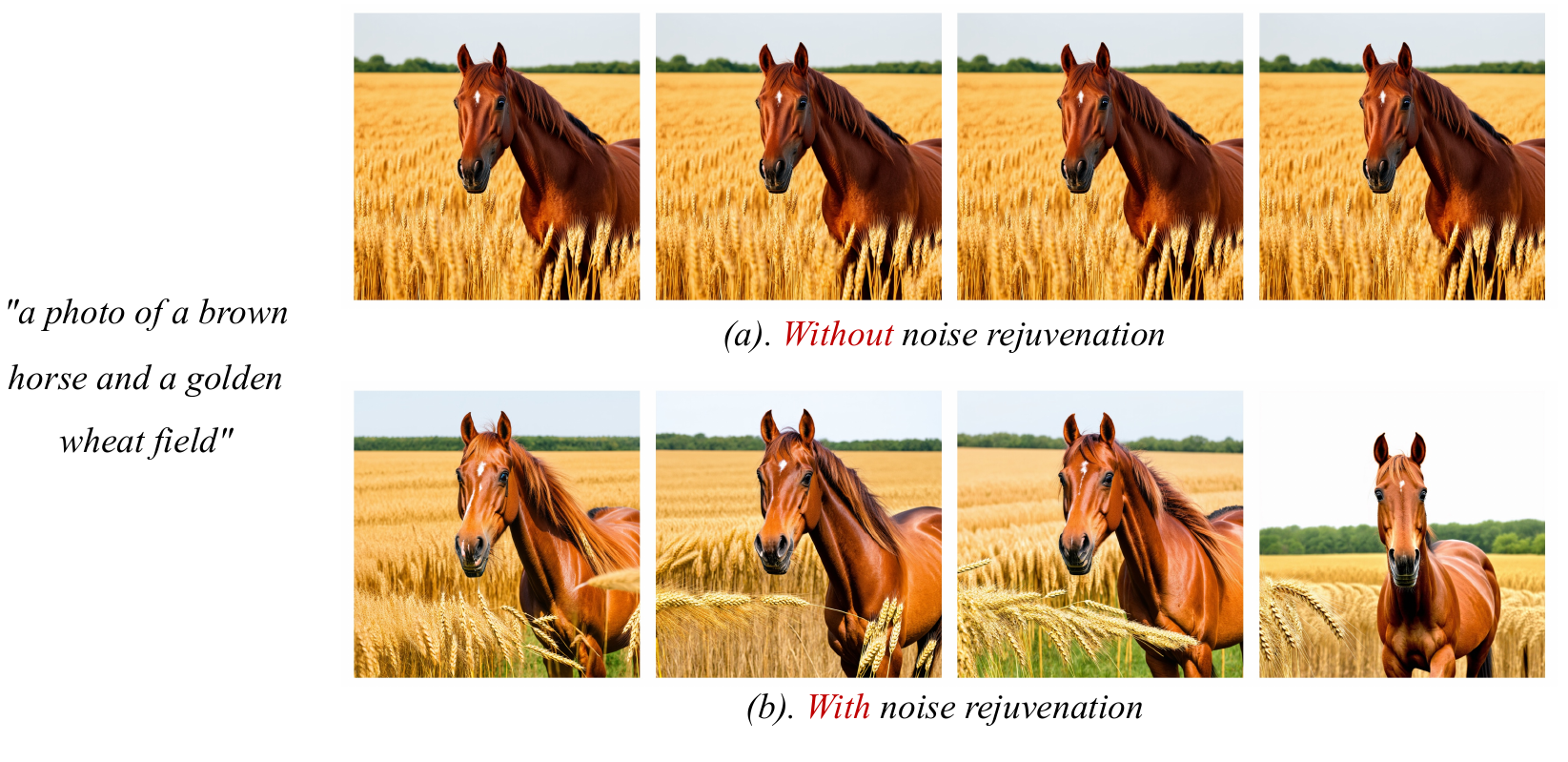}
\end{subfigure}
\begin{subfigure}{0.85\linewidth}
    \centering
    \setlength{\abovecaptionskip}{0.02in}
    \includegraphics[width=1.\linewidth]{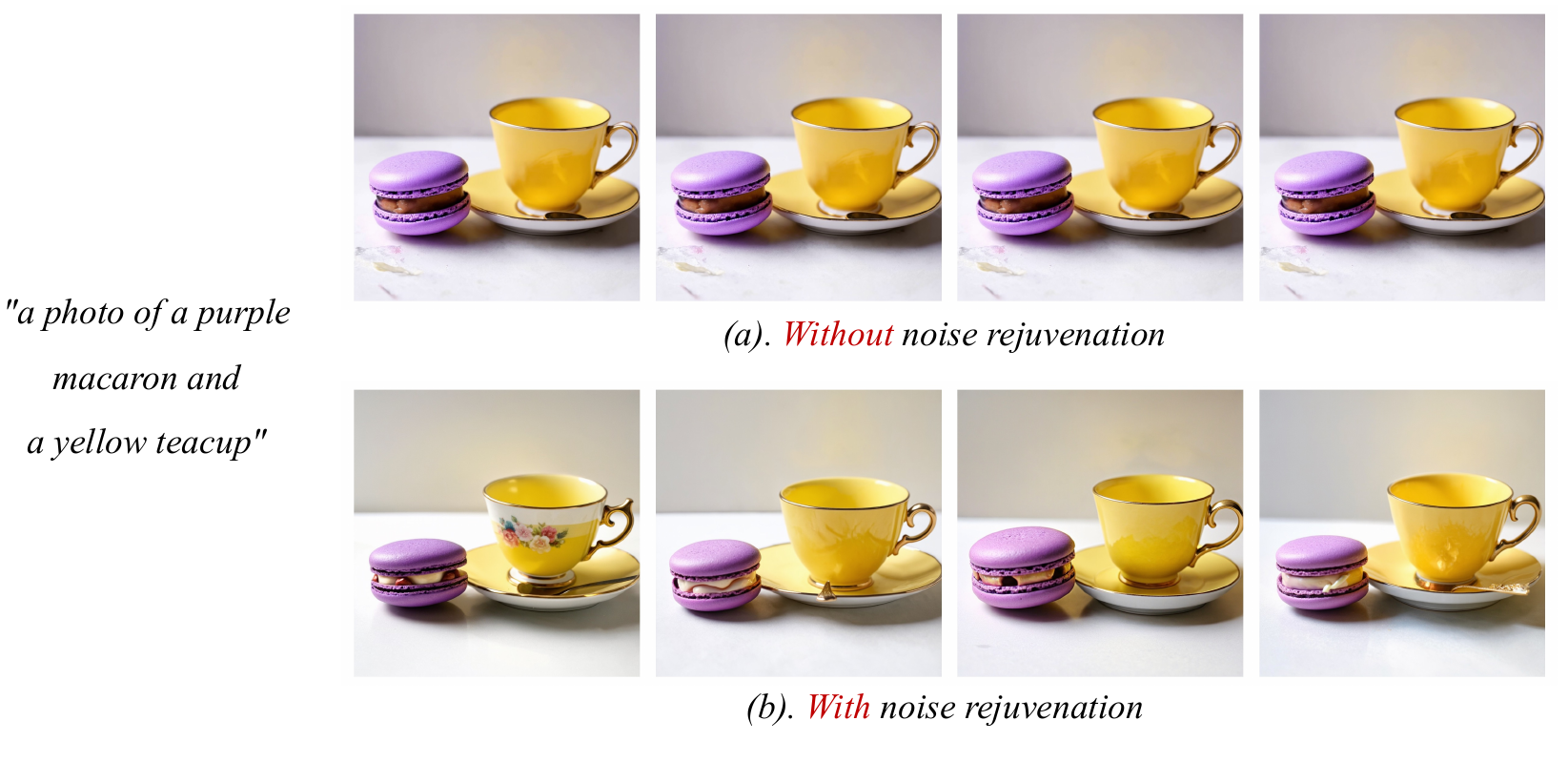}
\end{subfigure}
\begin{subfigure}{0.85\linewidth}
    \centering
    \setlength{\abovecaptionskip}{0.02in}
    \includegraphics[width=1.\linewidth]{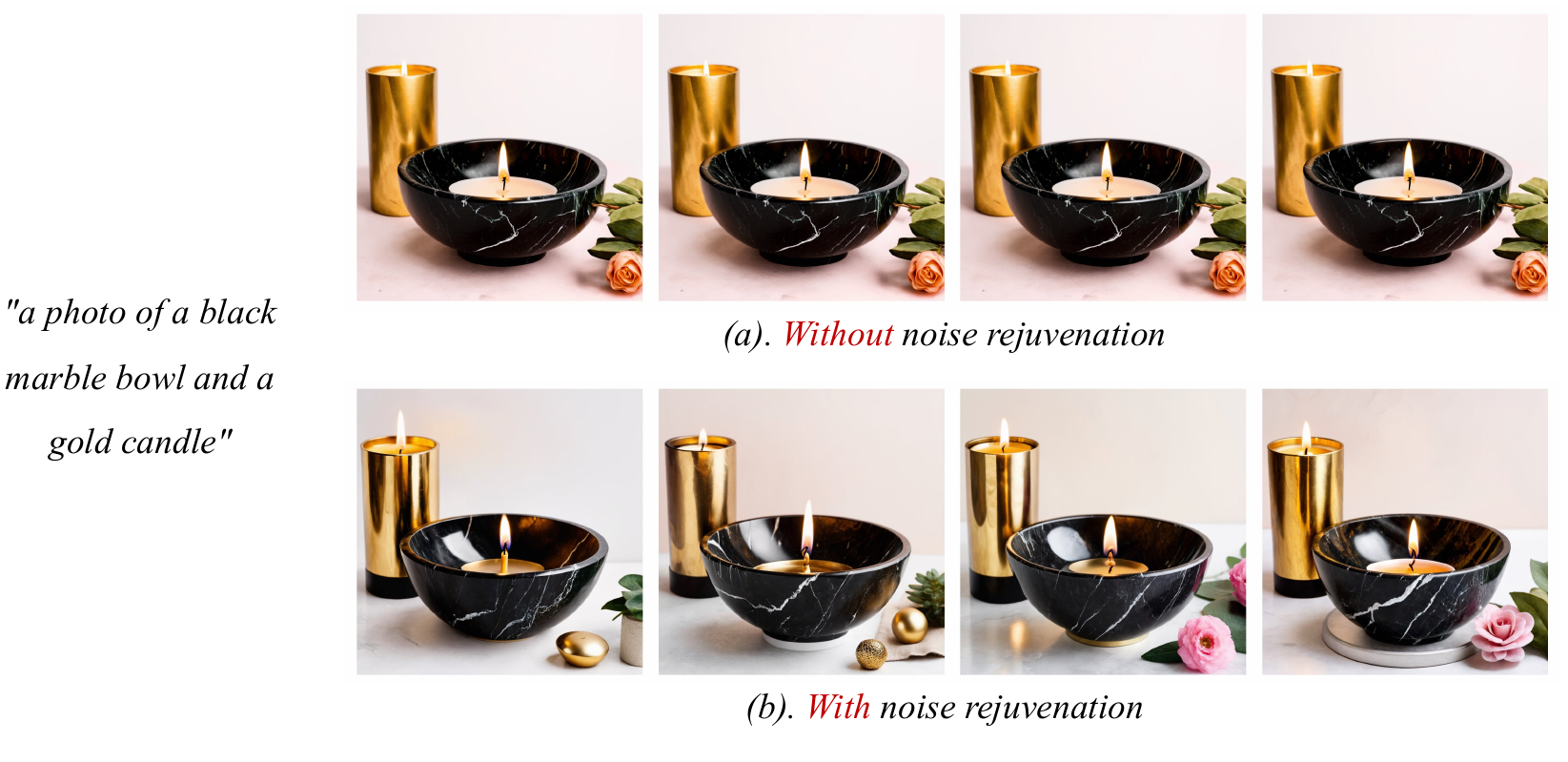}
\end{subfigure}
\caption{Additional examples comparing SMC on SD3 with and without noise rejuvenation. Without noise rejuvenation, resampled particles collapse to nearly identical outputs, whereas noise rejuvenation restores particle diversity. Please zoom in for details.}
\label{supp-fig:ode_collapse}
\end{figure}

As shown in \autoref{supp-fig:ode_collapse}, when all particles collapse to a single ancestor after resampling, our method is still able to explore diverse trajectories, resulting in images with distinct fine-grained details. In contrast, without noise rejuvenation, SMC degenerates to generating identical images. Please zoom in to view the local differences.
